\documentclass{article}

\PassOptionsToPackage{numbers, compress}{natbib}

\usepackage[final]{neurips_2021}

\usepackage[utf8]{inputenc} 
\usepackage[T1]{fontenc}    
\usepackage{microtype}
\usepackage{graphicx}
\usepackage{subcaption} 
\usepackage{booktabs} 

\usepackage{hyperref}
\usepackage{url}            
\usepackage{nicefrac}       
\usepackage{xcolor}         

\usepackage{amsfonts}
\usepackage{amsmath}
\usepackage{amssymb}
\usepackage{stmaryrd}
\usepackage{xcolor}
\usepackage{bbm}
\usepackage{enumitem}
\usepackage[normalem]{ulem}
\usepackage{wrapfig}

\newcommand{\R}{\mathbb{R}}
\newcommand{\A}{\mathcal{A}}

\newcommand{\argmax}{\operatorname*{\arg\max}}

\title{Program Synthesis Guided Reinforcement Learning for Partially Observed Environments}

\author{%
  Yichen David Yang\thanks{Correspondence to yicheny@csail.mit.edu} \\
  MIT EECS \& CSAIL\\
  \And
  Jeevana Priya Inala \\
  Microsoft Research\\
  \And
  Osbert Bastani \\
  University of Pennsylvania\\
  \And
  Yewen Pu \\
  Autodesk Research\\
  \And
  Armando Solar-Lezama \\
  MIT EECS \& CSAIL\\
  \And
  Martin Rinard \\
  MIT EECS \& CSAIL\\
}

\begin{document}

\maketitle

\begin{abstract}
A key challenge for reinforcement learning is solving long-horizon planning problems. Recent work has leveraged programs to guide reinforcement learning in these settings. However, these approaches impose a high manual burden on the user since they must provide a guiding program for every new task. Partially observed environments further complicate the programming task because the program must implement a strategy that correctly, and ideally optimally, handles every possible configuration of the hidden regions of the environment. We propose a new approach, \emph{model predictive program synthesis (MPPS)}, that uses program synthesis to automatically generate the guiding programs. It trains a generative model to predict the unobserved portions of the world, and then synthesizes a program based on samples from this model in a way that is robust to its uncertainty. In our experiments, we show that our approach significantly outperforms non-program-guided approaches on a set of challenging benchmarks, including a 2D Minecraft-inspired environment where the agent must complete a complex sequence of subtasks to achieve its goal, and achieves a similar performance as using handcrafted programs to guide the agent. Our results demonstrate that our approach can obtain the benefits of program-guided reinforcement learning without requiring the user to provide a new guiding program for every new task.
\end{abstract}

\section{Introduction}
\label{sec:introduction}

Reinforcement learning is a prominent technique for solving challenging planning and control problems~\cite{mnih2015human, arulkumaran2017deep}. Despite significant recent progress, solving long-horizon problems remains a significant challenge due to the combinatorial explosion of possible strategies. One promising approach to addressing these issues is to leverage \emph{programs} to guide the behavior of the agents~\cite{psketch, pagent, jothimurugan2019composable}. 
The approaches in this paradigm typically involve three key elements:
\begin{itemize}[topsep=0pt,itemsep=0ex,partopsep=1ex,parsep=1ex]
\item {\bf Domain-specific language (DSL):} For a given domain, the user defines a set of \emph{components} $c$ that correspond to intermediate subgoals that are useful for that domain (e.g., ``get wood'' or ``build bridge''), but leaves out how exactly to achieve these subgoals.
\item {\bf Task-specific program:} For every new task in the domain, the user provides a sequence of components (i.e. a program written in the DSL) that, if followed, enable the agent to achieve its goal in the task (e.g., [``get wood''; ``build bridge''; ``get gem'']).
\item {\bf Low-level neural policy:} For a given domain, the reinforcement learning algorithm learns an option~\cite{sutton1999between} that implements each component (i.e., achieves the subgoal specified by that component). Typically a neural policy is learned as each option.
\end{itemize}
Given a new task in a domain, the user provides a program in the DSL that describes a high-level strategy to solve that task. The agent then executes the program by deploying the sequence of learned options that correspond to the components in that program.

A key drawback of this approach is programming overhead: for every new task (a task consists of an instantiation of an environment and a goal), the user must analyze the environment, design a strategy to achieve the goal, and encode the strategy into a program, with a poorly written program producing a suboptimal agent. Furthermore, partially observed environments significantly complicate the programming task because the program must implement a strategy that correctly, and ideally optimally, handles every possible configuration of the hidden regions of the environment. 

To address this challenge, we propose a new approach, {\em model
predictive program synthesis (MPPS)}, that automatically synthesizes the guiding programs for program guided reinforcement learning.


MPPS works with a conditional generative model of the environment and a high level specification of the goal of the task to automatically synthesize a program that achieves the goal, with the synthesized program robust to uncertainty in the model. Because the automatically generated agent, and not the user, reasons about how to solve each new task, MPPS significantly reduces user burden. Given a goal specification $\phi$, the agent uses the following three steps to choose its actions:
\begin{itemize}[topsep=0pt,itemsep=0ex,partopsep=1ex,parsep=1ex]
\item {\bf Hallucinator:} First, inspired by world-models~\cite{world-models}, the agent keeps track of a conditional generative model $g$ over possible realizations of the unobserved portions of the environment.
\item {\bf Synthesizer:} 
Next,
the agent synthesizes a program $p$ that achieves $\phi$ assuming the hallucinator $g$ is accurate. Since world predictions are stochastic in nature, it samples multiple predicted worlds and computes the program that maximizes the probability of success.
\item {\bf Executor:} Finally, the agent executes the options corresponding to the components in the program $p=[c_1;...;c_k]$ for a fixed number of steps $N$.
\end{itemize}
If $\phi$ is not satisfied after $N$ steps, then the above process is repeated. Since the hallucinator now has more information (because the agent has explored more of the environment), the agent now has a better chance of achieving its goal. Importantly, the agent is implicitly encouraged to explore since it must do so to discover whether the current program can successfully achieve the goal $\phi$.

We instantiate our approach in the context of a 2D Minecraft-inspired environment~\cite{psketch, Sohn18, pagent}, which we call ``craft,'' and a ``box-world'' environment \cite{drrl}. We demonstrate that our approach significantly outperforms non-program-guided approaches, while achieving a similar performance as using handcrafted programs to guide the agent. In addition, we demonstrate that the policy we learn can be transferred to a continuous variant of the craft environment, where the agent is replaced by a MuJoCo~\cite{mujoco} Ant. Thus, our approach can obtain the benefits of program-guided reinforcement learning without requiring the user to provide a new guiding program for every new task.\footnote{The code is available at: \url{https://github.com/yycdavid/program-synthesis-guided-RL}}

\begin{figure*}
\centering
\begin{subfigure}[]{0.08\textwidth}
\centering
\includegraphics[width=\textwidth]{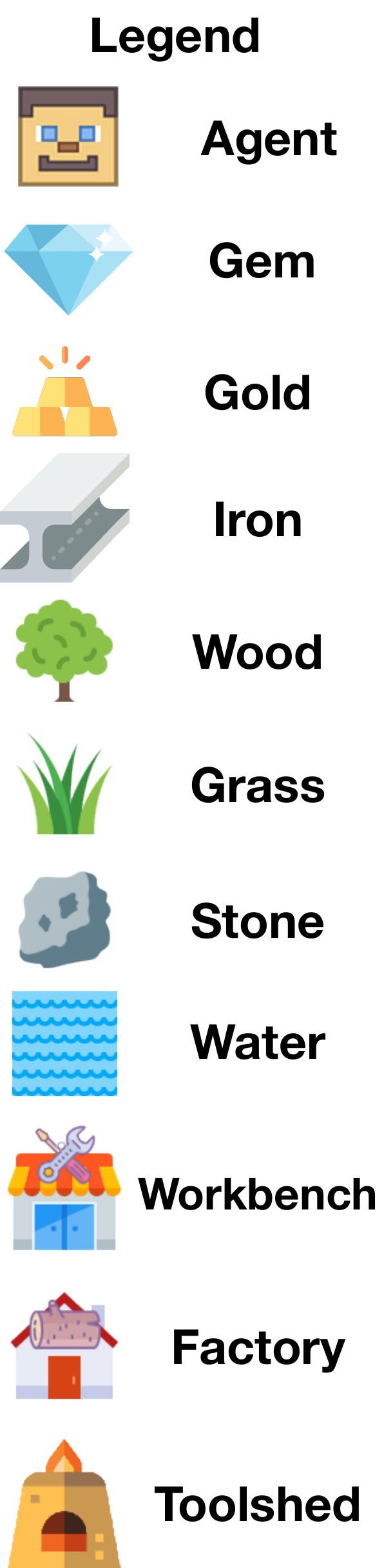}
\vspace{0.5em}
\end{subfigure}
\begin{subfigure}[]{0.44\textwidth}
\centering
\includegraphics[width=\textwidth]{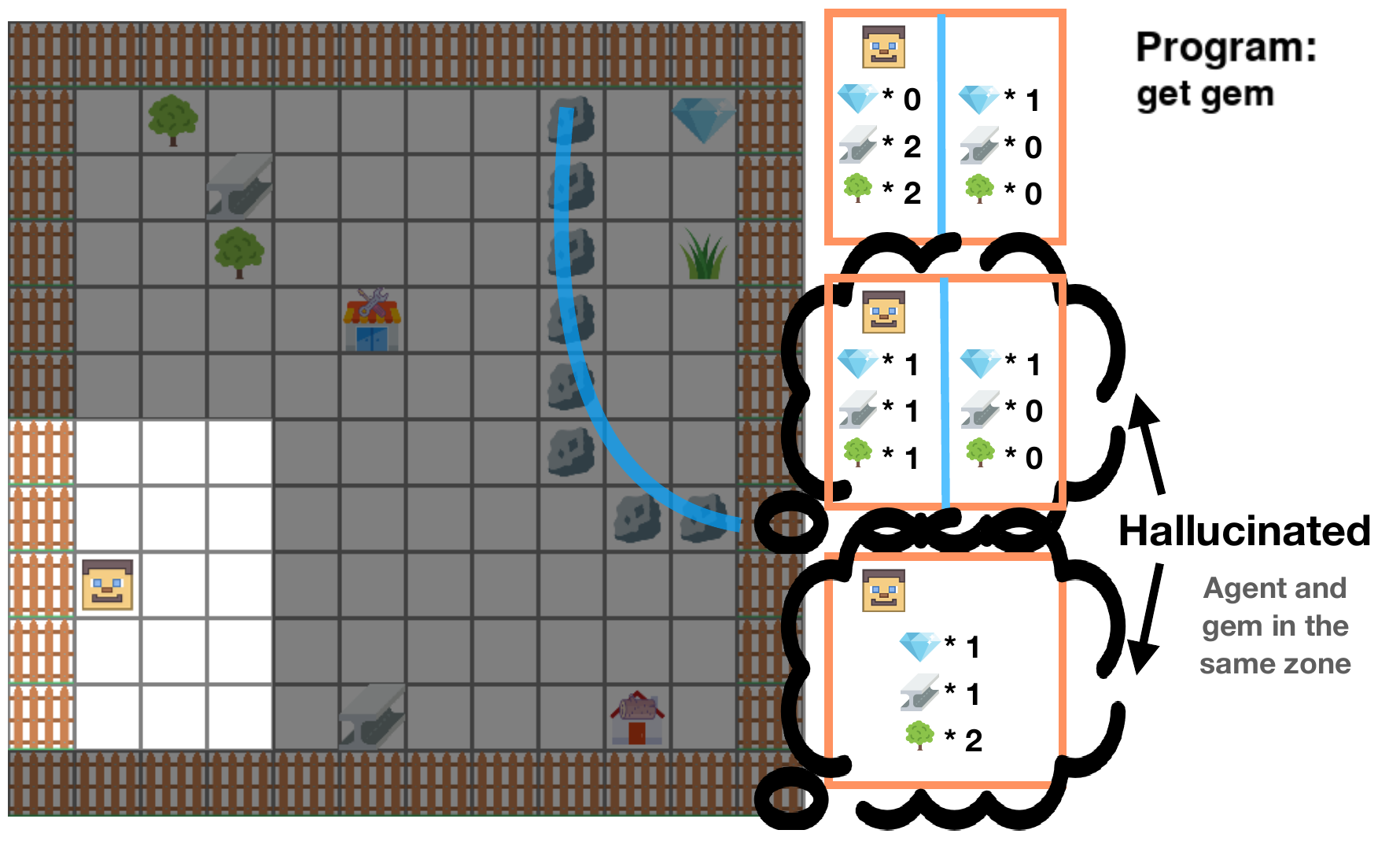}
\caption{}
\label{fig:map}
\end{subfigure}
\begin{subfigure}[]{0.44\textwidth}
\centering
\includegraphics[width=\textwidth]{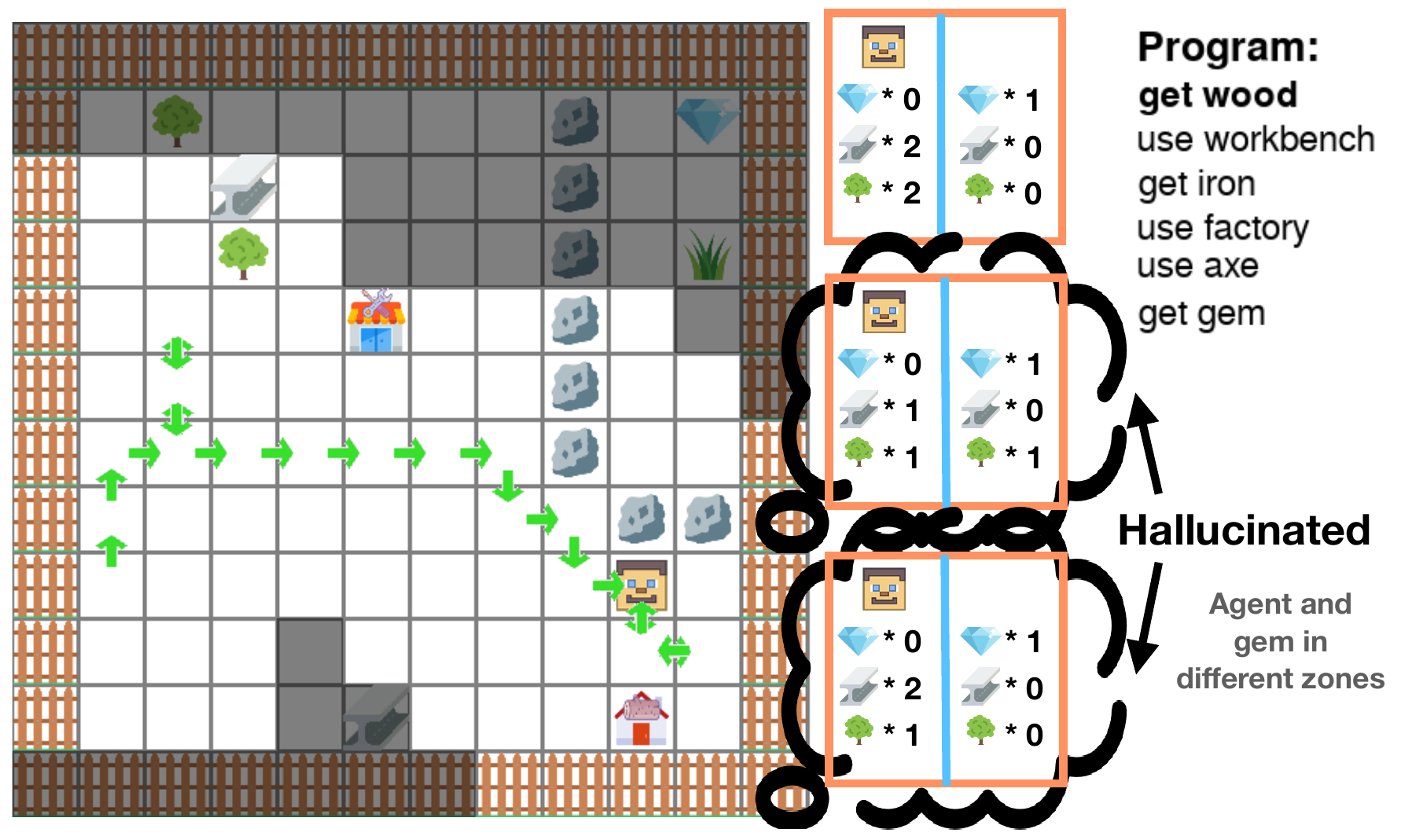}
\caption{}
\label{fig:replan}
\end{subfigure}
\caption{
(a) The initial state of an example task for the craft environment. Bright regions are observed and dark ones are unobserved. This particular map has two \emph{zones} separated by a stone boundary (blue line). The first zone contains the agent, 2 irons, and 2 woods; the second contains 1 grass and 1 gem (the goal). 
The agent represents the high-level structure of the map (e.g., resources in each zone) using state features. The ground truth features are in the top-right; we only show the counts of gems, irons, and woods in each zone and the zone containing the agent. The two thought bubbles below are features hallucinated by the agent based on the observed parts of the map. In both, the zone that the agent is in contains a gem, so the synthesized program is ``get gem''
(b) The state after the agent took 20 steps (green arrows), failed to obtain the gem, and is now re-synthesizing the program.
Having explored more of the map, it predicts that the gem is in a different zone, indicated by its two hallucinations. As a result, it synthesizes a program that includes building and using an axe to break the stone, which leads to successful completion of the task.
}
\end{figure*}

\textbf{Related work.} 
%
In general, program guidance makes reinforcement learning more tractable in at least two ways: (i) it provides intermediate rewards and (ii) it reduces the size of the search space of the policy by decomposing the policy into separate components. Previous research in program guided reinforcement learning demonstrates the benefits of this approach to guide reinforcement learning in the craft environment~\cite{pagent}. 
This previous research requires the user to provide both a DSL for the domain and a program for every new task. Furthermore, their approach requires that the user includes conditional statements in the program to handle partial observability, which imposes an even greater burden on the user. In contrast, we only require the user to provide a specification encoding the goal for each new task, and automatically handle partial observability.

There has been work enabling users to write specifications in a high-level language based on temporal logic \cite{jothimurugan2019composable}, with these specifications then translated into shaped rewards to guide learning. Furthermore, recent work has shown that even if the subgoal encoded by each component is omitted, the program (i.e., a sequence of symbols) can still aid learning~\cite{psketch}. Unlike our approach, this previous work requires the user to provide the guiding programs and does not handle partial observability. 

More broadly, our work fits into the literature on combining high-level planning with reinforcement learning. In particular, there is a long literature on planning with options~\cite{sutton1999between} (also known as \emph{skills}~\cite{hausman2018learning}), including work on inferring options~\cite{stolle2002learning}. 
Most of these approaches focus on MDPs with discrete state and action spaces and fully observed environments. Recent work~\cite{abel2020value,jothimurugan2021abstract,jothimurugan2021compositional,deepsynth,dac2019,wulfmeier2021,zhang2021hier,Bagaria2020Option,tessler2016deep,li2020subpolicy} addresses the challenge of handling continuous state and action spaces by combining high-level planning with reinforcement learning to handle low-level control, but does not handle the challenge of partial observations, whereas our work tackles both challenges.

Classical STRIPS planning~\cite{fikes1971strips} cannot handle uncertainty in the realization of the environment. Replanning~\cite{stentz1995focussed} can be used to handle small changes to an initially known environment, but cannot handle environments that are initially completely unknown. There has been work on hierarchical planning in POMDPs~\cite{charlin2007automated,toussaint2008hierarchical}, but this research does not incorporate predicate abstractions (i.e., state features) that can be used, for example, to handle continuous state and action spaces. 
Given multiple possible environments, generalized planning~\cite{genplan18,genplan19,genplanthesis,genplan11} can be used to compute a plan that is valid for all of them. However, in our setting, oftentimes no such plan exists. We instead synthesize a plan that is valid in a maximal number of hallucinated environments. 
There is also prior work on planning in partially observable environments~\cite{bonet1998, Draper1994}. Unlike our approach, these approaches assume that the effective state space is small, which enables them to compile the problem into a concrete POMDP which can be efficiently solved using POMDP algorithms.
We leverage program synthesis~\cite{sketch} with the world models approach~\cite{world-models} to address these issues; generally speaking, our solver-aided plan synthesis approach is more flexible than existing planning algorithms that target narrower problem settings.

Finally, there has broadly been recent interest in using program synthesis to learn programmatic policies that are more interpretable~\cite{verma2018programmatically,verma2019imitation,inala2021neurosymbolic}, verifiable~\cite{bastani2018verifiable,verma2019verifiable,anderson2020neurosymbolic}, and generalizable~\cite{inala2020synthesizing}. In contrast, we are not directly synthesizing the policy, but a program to guide the policy.
Appendix \ref{append:related} discusses additional related work in a broader context.

\section{Motivating Example}

Figure \ref{fig:map} shows a 2D Minecraft-inspired crafting game. In this grid world, the agent can navigate and collect resources (e.g., wood), build tools (e.g., a bridge) at workshops using collected resources, and use the tools to traverse obstacles (e.g., use a bridge to cross water). The agent can only observe the $5\times 5$ grid around its current position; since the environment is static, any previously observed cells remain visible. A single task consists of a randomly generated map (i.e., the environment) and goal (i.e., obtain a certain resource or build a certain tool).
We consider the meta-learning setting~\cite{finn2017model}: we have a set of training tasks for learning the policy, and our goal is to have a policy that works well on new tasks occurring in the future.

\begin{figure}
\centering
\begin{subfigure}[]{0.44\textwidth}
\centering
\input{dsl}
\caption{}
\label{fig:dsl}
\end{subfigure}
\begin{subfigure}[]{0.55\textwidth}
\centering
\includegraphics[width=\textwidth]{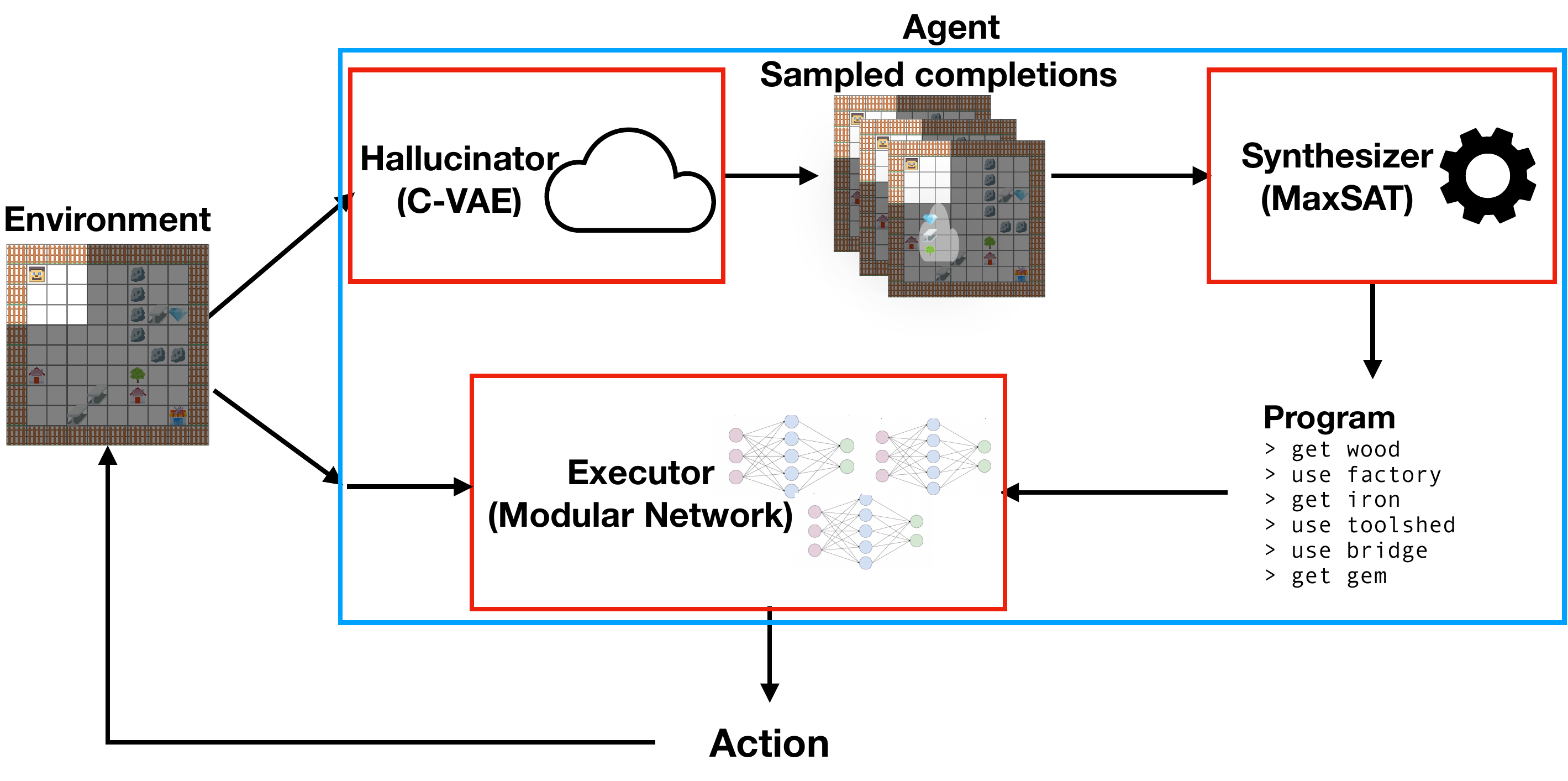}
\caption{}
\label{fig:schematic}
\end{subfigure}
\caption{
(a) DSL of components for the craft environment; the three kinds of components are get resource ($R$), use tool ($T$), and use workshop ($W$).
(b) Architecture of our agent (the blue box).}
\end{figure}

\textbf{DSL.}
A premise of our approach is a user-provided DSL consisting of components useful for the domain. Figure~\ref{fig:dsl} shows the DSL for the craft environment. For each component, the user also specifies what the component is expected to achieve as a logical predicate.
To deal with high-dimensional state spaces, the logical predicates are expressed over features $\alpha(s)$ of the state---e.g.,
the logical predicate for ``get wood'' is
\begin{align*}
\forall i, j\;.\; & ( z^- = i \wedge z^+ = j ) \Rightarrow ( b_{i,j}^- = \text{connected} )
\wedge ( \rho^+_{j,\text{wood}} = \rho_{j,\text{wood}}^- - 1 ) \wedge ( \iota^+_{\text{wood}} = \iota_{\text{wood}}^- + 1 ).
\end{align*}
This predicate is over two sets of features: (i) features $\alpha(s^-)$, denoted by a $-$, of the initial state $s^-$ (i.e., where execution of the component starts), and (ii) features $\alpha(s^+)$, denoted by a $+$, of the final state $s^+$ (i.e., where the subgoal is achieved and execution of the component terminates). The first feature is the categorical feature $z$ that indicates the zone containing the agent. In particular, we divide the map into zones that are regions separated by obstacles such as water and stone---e.g., the map in Figure \ref{fig:map} has two zones: (i) the region containing the agent, and (ii) the region blocked off by stones. Now, the feature $b_{i,j}$ indicates whether zones $i$ and $j$ are connected, $\rho_{i,r}$ denotes the count of resource $r$ in zone $i$, and $\iota_r$ denotes the count of resource $r$ in the agent's inventory.

Thus, this formula says that (i) the agent goes from zone $i$ to $j$, (ii) $i$ and $j$ are connected, (iii) the count of wood in the agent's inventory increases by one, and (iv) the count of wood in zone $j$ decreases by one. Appendix \ref{append:semantics} describes the full set of components we use.

\textbf{Approach.}
Before solving any new tasks, for each component $c$, we use reinforcement learning to train an option $\tilde{c}$ that attempts to achieve the subgoal encoded by $c$. 
Given a new task, the user specifies the goal of the task as a logical predicate $\phi$. Encoding the goal is typically simple; for example, the goal of the task in Figure \ref{fig:map} is getting gem, which is encoded as $\phi:= \iota_{\textrm{gem}}\geq 1$. Then the agent attempts to solve the task as follows.

First, based on the observations so far, the agent uses a hallucinator $g$ to predict multiple potential worlds, each of which represents a possible realization of the full map. Rather than predicting concrete states, it suffices to predict the state features. For instance, Figure \ref{fig:map} shows two samples of the world predicted by $g$; here, the only values it predicts are the number of zones in the map, the type of the boundary between the zones, and the counts of the resources and workshops in each zone. In this example, the first predicted world contains two zones, and the second contains one zone. Note that in both predicted worlds, there is a gem located in same zone as the agent.

Next, the agent synthesizes a program $p$ that achieves the goal $\phi$ in a maximal number of predicted worlds. The synthesized program in Figure \ref{fig:map} is a single component ``get gem,'' which refers to searching the current zone (or zones already connected with the current zone) for a gem. Note that this program achieves the goal for the predicted worlds shown in Figure \ref{fig:map}.

Finally, the agent executes the program $p=[c_1;...;c_k]$ for a fixed number $N$ of steps. In particular, it executes the policy $\pi_\tau$ of option $\tilde{c}_\tau=(\pi_\tau,\beta_\tau)$ corresponding to $c_\tau$ until the termination condition $\beta_\tau$ holds, upon which it switches to executing $\pi_{\tau+1}$. In our example, there is only one component ``get gem,'' so it executes the policy for this component until the agent finds a gem.

In this case, the agent fails to achieve its goal $\phi$ since there is no gem in its current zone. Thus, it repeats the above process. Since it now has more observations, $g$ more accurately predicts the world---e.g., Figure \ref{fig:replan} shows the intermediate step when the agent re-plans. Note that it now correctly predicts that the only gem is in the second zone. Thus, the newly synthesized program is
\begin{align*}
p=&[\underbrace{\text{get wood};\text{use workbench};\text{get iron};\text{use factory};}_{\text{for building axe}}\;
\text{use axe};\text{get gem}].
\end{align*}
That is, it builds an axe to break the stone so it can get to the zone containing the gem.
Finally, the agent executes this new program, which successfully finds the gem.

\section{Problem Formulation}
\label{sec:prelmiinaries}

\textbf{POMDP.}
We consider a partially observed Markov decision process (POMDP)
with states $\mathcal{S}\subseteq \R^n$,
actions $\A\subseteq\R^m$,
observations $\mathcal{O}\subseteq\R^q$,
initial state distribution $\mathcal{P}_0$, 
observation function $h: \mathcal{S} \rightarrow \mathcal{O}$,
and transition function $f: \mathcal{S} \times \mathcal{A} \rightarrow \mathcal{S}$.
Given initial state $s_0 \sim \mathcal{P}_0$, policy $\pi:\mathcal{O}\to\mathcal{A}$, and time horizon $T\in\mathbb{N}$, the generated trajectory is $(s_0, a_0, s_1, a_1, \dots, s_T, a_T)$, where $o_t = h(s_t)$, $a_t = \pi(o_t)$, and $s_{t+1} = f(s_t, a_t)$.
We assume the state includes the unobserved parts of the environment---e.g.,
in the craft environment, it represents both the entire map and the agent's current position and inventory.

We consider a meta-learning setting, where we have a set of sampled training tasks (world configurations and goal) and a set of test tasks. Our goal is to learn a policy using the training set that achieves good performance on the test set.

\textbf{User-provided components.}
We consider programs $p=[c_1;...;c_k]$ composed of \emph{components} $c_\tau\in C$. We assume the user provides the set of components $C$ that are useful for the domain. Importantly, these components only need to be provided once for a domain; they are shared across all tasks in this domain.
Each component is specified as a logical predicate that encodes the intended behavior of that component.
More precisely, $c$ is a logical predicate over $s^-$ and $s^+$, where $s^-$ denotes the initial state before executing $c$ and $s^+$ denotes the final state after executing $c$. For instance, the component
\begin{align*}
c\equiv(s^-=s_0\Rightarrow s^+=s_1)\wedge(s^-=s_2\Rightarrow s^+=s_3)
\end{align*}
says that if the POMDP is currently in state $s_0$, then $c$ should transition it to $s_1$, and if it is currently in state $s_2$, then $c$ should transition it to $s_3$.
Rather than defining $c$ over the concrete states, we can define it over features $\alpha(s^-)$ and $\alpha(s^+)$ of the states in order to handle high-dimensional state spaces.

\textbf{User-provided goal specification.}
The goal of each task is specified with a logical predicate $\phi$ over the final state; as with components, $\phi$ may be specified over features $\alpha(s)$ instead of concrete states. Our objective is to design an agent that can achieve any given specification $\phi$ (i.e., act in the POMDP to reach a state that satisfies $\phi$) as quickly as possible.

\section{Model Predictive Program Synthesis}
\label{sec:method}

We describe here the architecture of our agent, depicted in Figure \ref{fig:schematic}. It is composed of three parts: the \textit{hallucinator} $g$, which predicts possible worlds; the \textit{synthesizer}, which generates a program $p$ that maximizes the probability of success according to worlds sampled from $g$; and the \textit{executor}, which follows $p$ to act in the POMDP. These parts are run once every $N$ steps to generate a program $p$ to execute for the subsequent $N$ steps, until the user-provided specification $\phi$ is achieved.


\textbf{Hallucinator.}
First, the hallucinator is a conditional generative model trained to predict the unobserved parts of the environment given the observations. 
To be precise, the hallucinator $g$ encodes a distribution $g(s\mid o)$, which is trained to approximate the actual distribution $P(s\mid o)$. 
Then, at each iteration (i.e., once every $N$ steps), our agent samples $m$ worlds $\hat{s}_1,...,\hat{s}_m\sim g(\cdot\mid o)$. Our technique can work with any type of conditional generative model as the hallucinator; in our experiments, we use a conditional variational auto-encoder (CVAE) \cite{cvae}.

When using state features, we can have $g$ directly predict the features; this approach works since the synthesizer only needs to know the values of the features to generate a program (see below).

\textbf{Synthesizer.}
The synthesizer computes a program that maximizes the probability of satisfying $\phi$:
\begin{align}
p^*&=\argmax_p\mathbb{E}_{P(s\mid o)}\mathbbm{1}[p\text{ solves }\phi\text{ for }s]
\approx\argmax_p\frac{1}{m}\sum_{j=1}^m\mathbbm{1}[p\text{ solves }\phi\text{ for }\hat{s}_j], \label{eqn:monte-carlo}
\end{align}
where the $\hat{s}_j$ are samples from $g$.
The objective (\ref{eqn:monte-carlo}) can be expressed as a MaxSAT problem~\cite{maxsat}. In particular, suppose for now that we are searching over programs $p=[c_1;...;c_k]$ of fixed length $k$. Then, consider the constrained optimization problem
\begin{align}
\label{eqn:maxsat}
\argmax_{\xi_1,...,\xi_k}\frac{1}{m}\sum_{j=1}^m\exists s_1^-,s_1^+,...,s_k^-,s_k^+\;.\;\psi_j,
\end{align}
where $\xi_\tau$ and $s_\tau^\delta$ (for $\tau\in\{1,...,k\}$ and $\delta\in\{-,+\}$) are the optimization variables. Here, $\xi_1,...,\xi_k$ encodes the program $p=[c_1;...;c_k]$, and $\psi_j$ encodes the constraints that $p$ solves $\phi$ for world $\hat{s}_j$---i.e.,
\begin{align*}
\psi_j\equiv\psi_{j,\text{start}}\wedge\left[\bigwedge_{\tau=1}^k\psi_{j,\tau}\right]\wedge\left[\bigwedge_{\tau=1}^{k-1}\psi_{j,\tau}'\right]\wedge\psi_{j,\text{end}},
\end{align*}
where
(i) $\psi_{j,\text{start}}\equiv(s_1^-=\hat{s}_j)$
encodes that the initial state is $\hat{s}_j$,
(ii) $\psi_{j,\tau}\equiv\big((\xi_\tau=c)\Rightarrow c(s_\tau^-,s_\tau^+)\big)$
encodes that if the the $\tau$th component of $p$ is $c_\tau=c$, then the transition from $s_\tau^-$ to $s_\tau^+$ on step $\tau$ satisfies $c(s_\tau^-,s_\tau^+)$,
(iii) $\psi_{j,\tau}'\equiv(s_\tau^+=s_{\tau+1}^-)$
encodes that the final state of the $\tau$th step equals the initial state the $(\tau+1)$th step, and
(iv) $\psi_{j,\text{end}}\equiv\phi(s_j^+)$
encodes that the final state of the last component should satisfy the user-provided goal $\phi$. We use a MaxSAT solver to solve (\ref{eqn:maxsat})~\cite{z3}. Given a solution $\xi_1=c_1,...,\xi_k=c_k$, the synthesizer returns the corresponding program $p=[c_1;...;c_k]$.

We incrementally search for longer and longer programs, starting from $k=1$ and incrementing $k$ until either we find a program that achieves at least a minimum objective value, or we reach a maximum program length $k_{\textrm{max}}$, at which point we use the best program found so far.

\textbf{Executor.}
For each user-provided component $c\in C$, we use reinforcement learning to learn an option $\tilde{c}=(\pi,\beta)$ that executes the component, where $\pi:\mathcal{O}\to\mathcal{A}$ is a policy and $\beta:\mathcal{O}\to\{0,1\}$ is a termination condition. 
The executor runs the synthesized program $p=[c_1;...;c_k]$ by deploying each corresponding option 
$\tilde{c}_\tau=(\pi_\tau,\beta_\tau)$ in sequence, starting from $\tau=1$. In particular, it uses action $a_t=\pi_\tau(o_t)$ at each time step $t$, where $o_t$ is the observation on that step, until $\beta_\tau(o_t)=1$, at which point it increments $\tau\gets\tau+1$. It continues until either it has completed running the program ($\beta_k(o_t)=1$), or after $N$ steps. In the former case, by construction, the goal $\phi$ has been achieved, so the agent terminates. In the latter case, the agent iteratively reruns the hallucinator and the synthesizer based on the current observation to get a new program. At this point, the hallucinator likely has additional information about the environment, so the new program has a greater chance of success.

\section{Learning Algorithm}
\label{sec:learning}

Next, we describe our algorithm for learning the parameters of models used by our agent. In particular, there are two parts that need to be learned: (i) the parameters of the hallucinator $g$ and (ii) the options $\tilde{c}$ based on the user-provided components $c$.

\textbf{Hallucinator.}
The goal is to train the hallucinator $g(s\mid o)$ to approximate the actual distribution $P(s\mid o)$ of the state $s$ given the observation $o$. We obtain samples ${(o_t, s_t)}$ from the training tasks using rollouts from a random agent and train $g_{\theta}(s\mid o)$ using supervised learning. In our experiments, we take $g_{\theta}$ to be a CVAE and train it using the evidence lower bound (ELBo) on the log likelihood~\cite{kingma2013auto}.

\textbf{Executor.}
Our framework uses reinforcement learning to learn options $\tilde{c}$ that implement the user-provided components $c$; these options can be shared across multiple tasks. We use neural module networks \cite{psketch} as the model for the executor policy; but in general our approach can also work with other types of models. In particular, we take $\tilde{c}=(\pi,\beta)$, where $\pi:\mathcal{O}\to\mathcal{A}$ is a neural module and $\beta:\mathcal{O}\to\{0,1\}$ checks when to terminate execution. First, $\beta$ is constructed directly from $c$---i.e., it returns whether $c$ is satisfied based on the current observation $o$.
Next, we train $\pi$ on the training tasks, which consist of randomly generated initial states $s$ and goal specifications $\phi$. Just for training, we use the ground truth program $p$ synthesized based on the fully observed environment; this approach avoids the need to run the synthesizer repeatedly during training. Given $p$, we sample a rollout $\{(o_1, a_1, r_1),...,(o_T, a_T, r_T)\}$ by running the current options $c_\tau=(\pi_\tau,\beta_\tau)$ according to the order specified by $p$ (where $\pi_\tau$ is randomly initialized). We give the agent a reward $\tilde{r}$ at each step when it achieves the subgoal of the component $c_{\tau}$, as well as a final reward when it achieves the final goal $\phi$. Then, we use actor-critic reinforcement learning~\cite{ac} to update $\pi$.
Finally, we use curriculum learning to speed up training---i.e., we train using tasks that can be solved with shorter programs first~\cite{psketch}.

\section{Experiments}
\label{sec:exp}

We empirically show that our approach significantly outperforms prior approaches that do not leverage programs, and furthermore achieves similar performance as an oracle given the ground truth program.

\begin{figure}
\centering
\begin{subfigure}[]{0.21\textwidth}
\centering
\includegraphics[width=\textwidth]{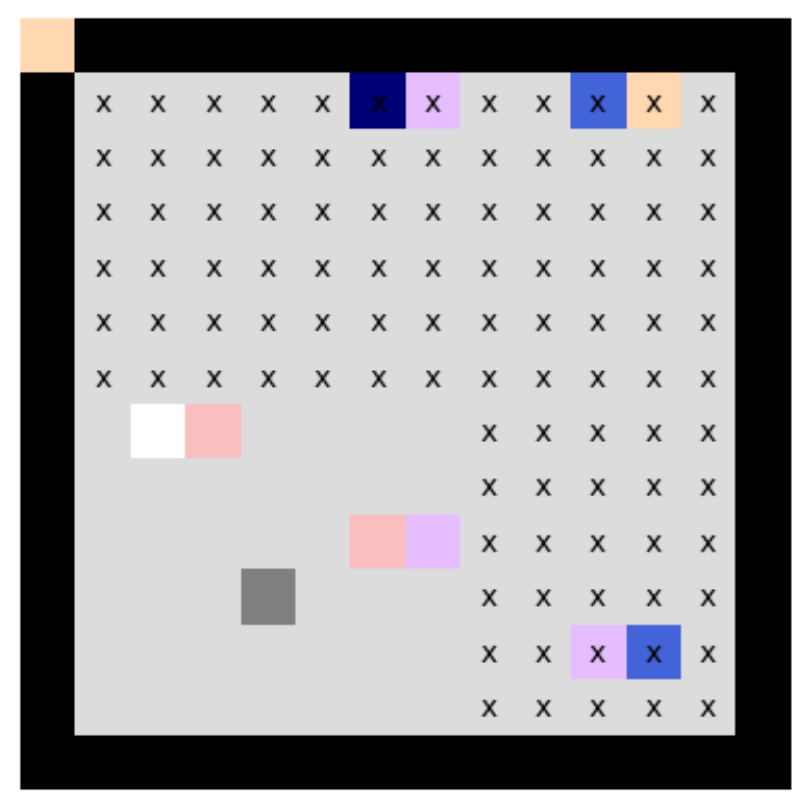}
\caption{}
\label{fig:box}
\end{subfigure}
\begin{subfigure}[]{0.2\textwidth}
\centering
\includegraphics[width=\textwidth]{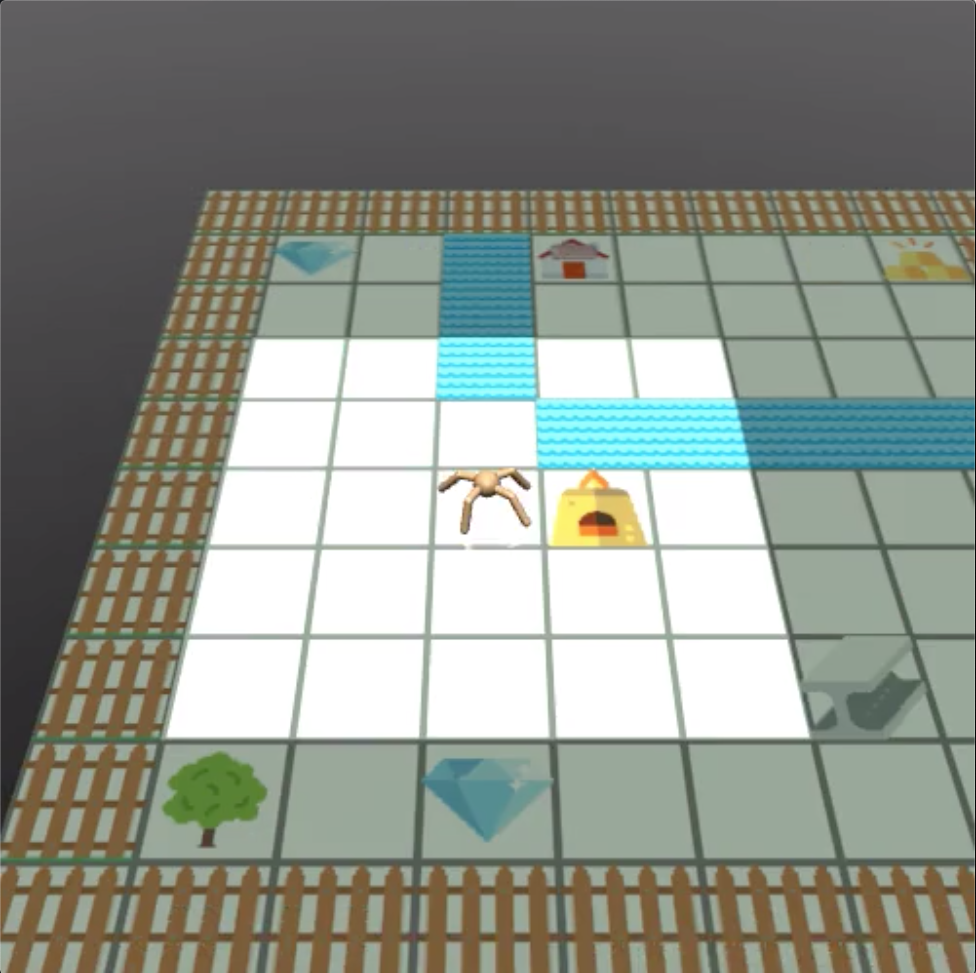}
\caption{}
\label{fig:ant-craft}
\end{subfigure}
\begin{subfigure}[]{0.26\textwidth}
\centering
\includegraphics[width=\textwidth]{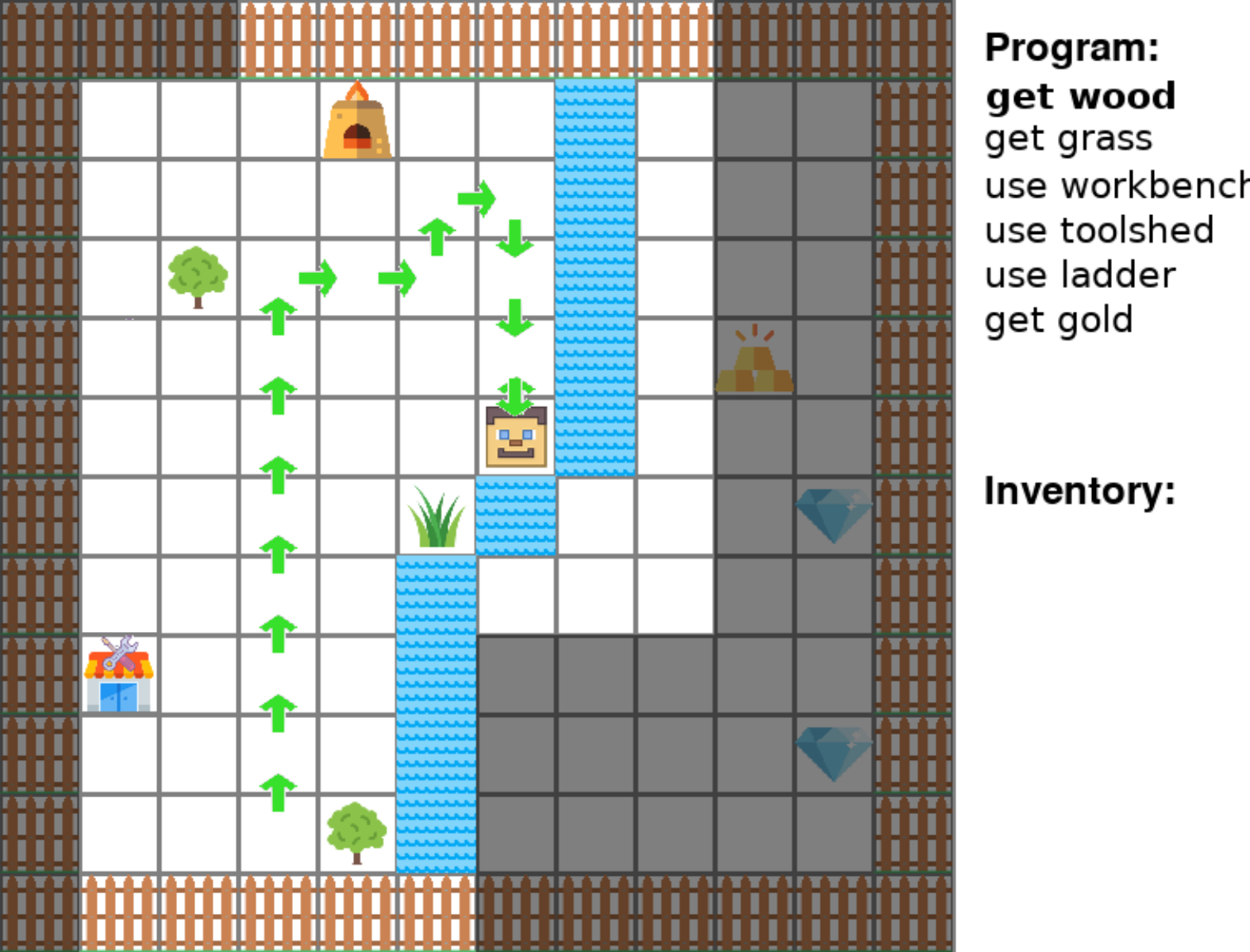}
\caption{}
\label{fig:ablation-optim}
\end{subfigure}
\begin{subfigure}[]{0.26\textwidth}
\centering
\includegraphics[width=\textwidth]{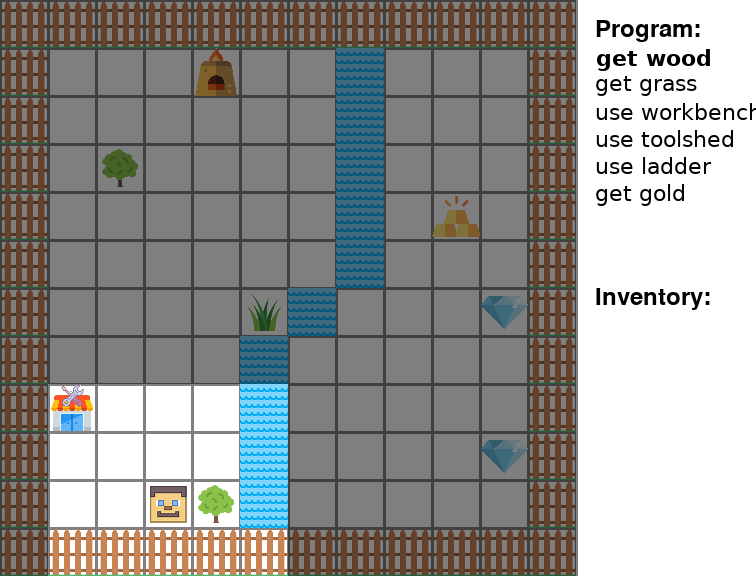}
\caption{}
\label{fig:ablation-ours}
\end{subfigure}
\caption{
(a) The box-world environment. The grey pixel denotes the agent. The goal is to get the white key. The unobserved parts of the map is marked with ``x''. The key currently held by the agent is shown in the top-left corner. In this map, the number of boxes in the path to the goal is 4, and it contains 1 distractor branch.
(b) The ant-craft environment. The policy needs to control the ant to perform the crafting tasks. 
(c,d) Comparison of behaviors between the optimistic approach (left) and our MPPS approach (right), in a task where the goal is to get gold. (c) The state when the optimistic approach first synthesizes the correct program instead of the (incorrect) one ``get gold''. It only does so after observing all the squares in its current zone. (d) The initial state of our MPPS strategy. It directly synthesizes the correct program, since the hallucinator knows the gold is most likely in the other zone based on the observations. Thus, the agent completes the task much more quickly. }
\end{figure}

\subsection{Benchmarks}

\textbf{2D-craft.}
We consider a 2D Minecraft-inspired game~\citep{psketch} (Figure \ref{fig:map}). A map is a $10\times10$ grid, where each grid cell is either empty or contains a resource (e.g., wood), obstacle (e.g., water), or workshop. 
Each task consists of a randomly sampled map, initial position, and goal (one of 10 possibilities, either getting a resource or building a tool), which typically require the agent to achieve several intermediate subgoals.
In contrast to prior work, our agent does not initially observe the entire map; instead, they can only observe cells within two units. Since the environment is static, any previously observed cells remain visible. The actions are discrete: moving in one of the four directions, picking up a resource, using a workshop, or using a tool. The maximum episode length is $T=100$. 

\textbf{Box-world.}
Next, we consider box-world \cite{drrl}, which requires abstract reasoning. It is a $12\times12$ grid world with locks and boxes (Figure \ref{fig:box}). The agent is given a key to get started, and its goal is to unlock a white box. Each lock locks a box in the adjacent cell containing a key.
Lock and boxes are colored; the key needed to open a lock is in the box of the same color. The actions are to move in one of the four directions; the agent opens a lock and obtains the key simply by walking over it. We assume that the agent can unlock multiple locks with each key. The agent can only observe grid cells within a distance of 3 (as well as the previously observed cells).
Each task consists of a randomly sampled map and initial position, where the number of boxes in the path to the goal is randomly chosen between 1 to 4, and the number of ``distractor branches'' (i.e., boxes that the agent can open but does not help them reach the goal) is also randomly chosen between 1 to 4.

More details about the environments are described in Appendix \ref{append:env}

\subsection{Baselines}


\textbf{End-to-end.}
A set of DNN policies that solves the tasks end-to-end. It uses one DNN policy per type of goal, i.e. one network will be used to solve all tasks with the goal of ``get gem'', another network for tasks with the goal of ``build bridge''. This baseline is trained using the same actor-critic algorithm and curriculum learning strategy as described in Section~\ref{sec:learning}.

\textbf{World models~\cite{world-models}.}
This approach handles partial observability by using a generative model to predict the future.
It trains a VAE model that encodes the current observation $o_t$ into a latent vector $z_t$, and trains a recurrent model to predict $z_{t+1}$ based on $z_1,...,z_t$. Then, it trains a policy using the latent vectors from the VAE model and the recurrent model as inputs.

\textbf{Relational reinforcement learning~\cite{drrl}.}
For box-world, we also compare with this approach, which uses a relational module based on the multi-head attention mechanism \cite{transformer} for the policy network to facilitate relational reasoning.

\textbf{Oracle.}
Finally, we compare to an oracle, which is our approach but given the ground truth program (i.e., guaranteed to achieve $\phi$). This can be seen as the program-guided agent approach~\cite{pagent}. This baseline is an oracle since it strictly requires more information as input from the user.

\begin{figure*}
\centering
\begin{subfigure}[]{0.22\textwidth}
\centering
\includegraphics[width=\textwidth]{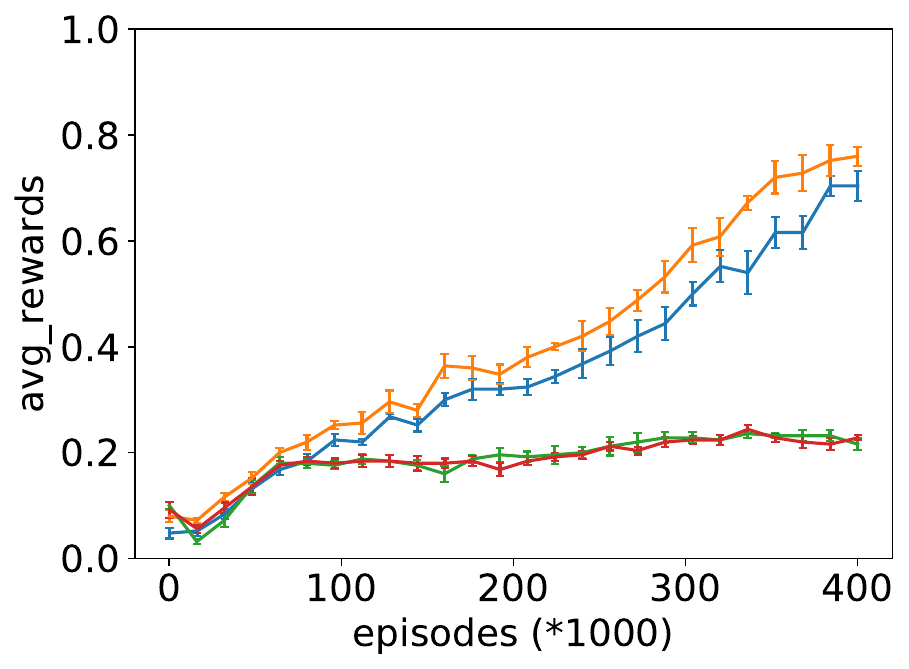}
\caption{}
\label{fig:reward-curve}
\end{subfigure}
\begin{subfigure}[]{0.22\textwidth}
\centering
\includegraphics[width=\textwidth]{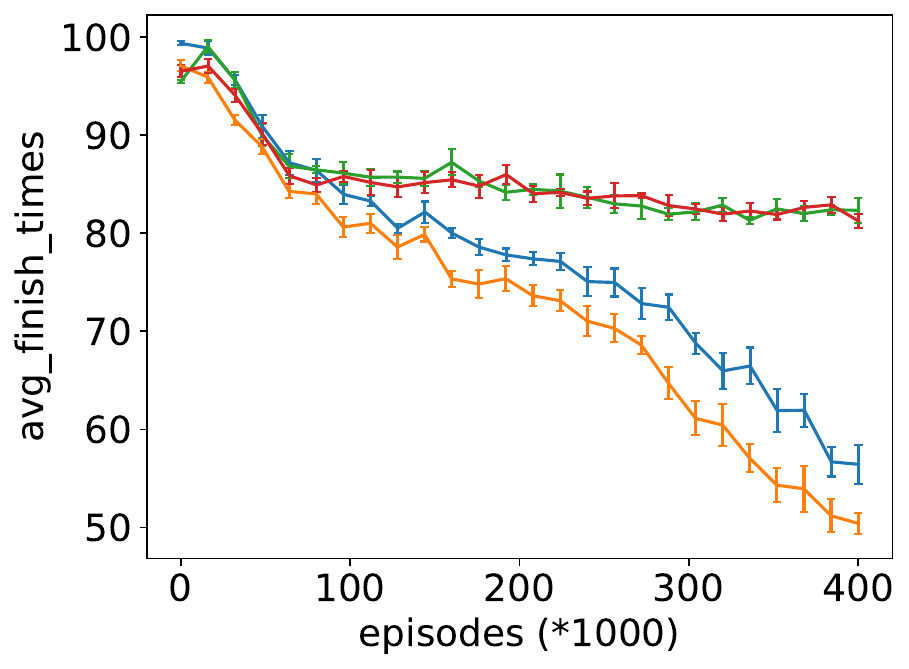}
\caption{}
\label{fig:time-curve}
\end{subfigure}
\begin{subfigure}[]{0.22\textwidth}
\centering
\includegraphics[width=\textwidth]{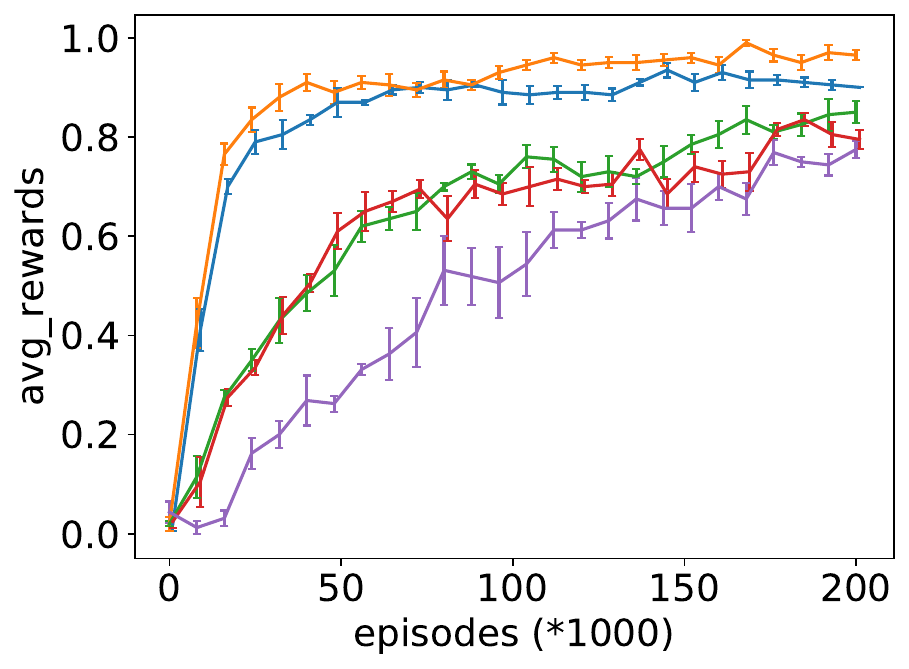}
\caption{}
\label{fig:box-reward-curve}
\end{subfigure}
\begin{subfigure}[]{0.22\textwidth}
\centering
\includegraphics[width=\textwidth]{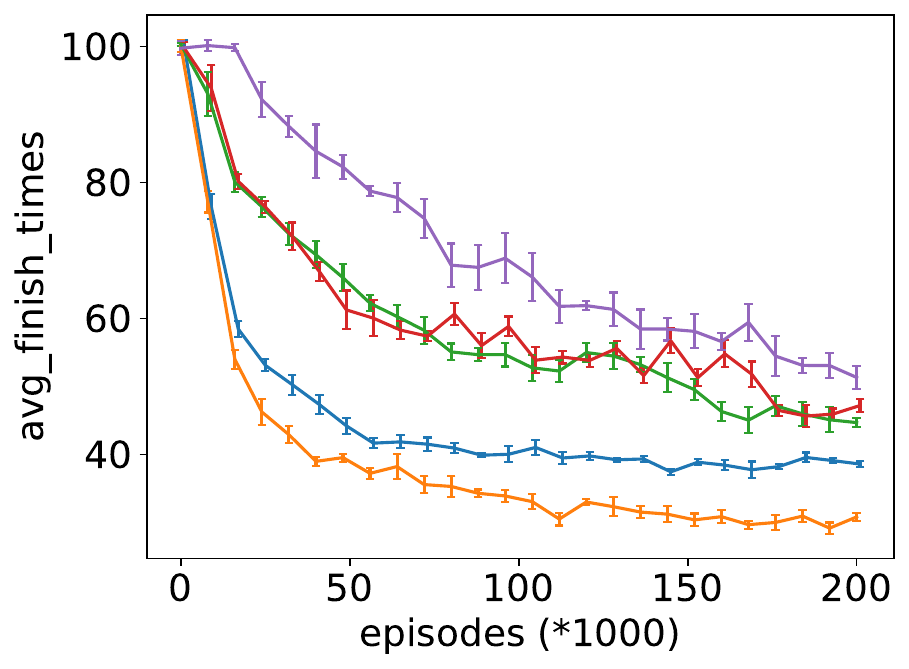}
\caption{}
\label{fig:box-finish-time}
\end{subfigure}
\begin{subfigure}[]{0.09\textwidth}
\centering
\includegraphics[width=\textwidth]{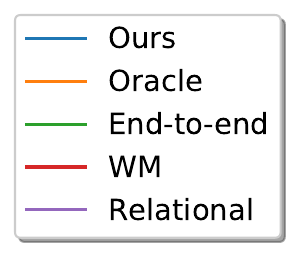}
\vspace{0.5em}
\end{subfigure}
\caption{ (a,b) Training curves for the 2D-craft environment.
(c,d) Training curves for the box-world environment.
(a,c) The average reward on the test set over the course of training; the agent gets a reward of 1 if it successfully finishes the task within the time horizon, and 0 otherwise. (b,d) The average number of steps taken to complete the tasks in the test set. We run all the training with 5 different random seeds, and report the mean and standard error of each metric. We show our approach (``Ours''), program guided agent (``Oracle''), end-to-end neural policy (``End-to-end''), world models (``WM''), and relational reinforcement learning (``Relational'').
For our approach, we include the episodes used for training the hallucinator in the starting parts of the training curve; since the number of episodes used for hallucinator training is substantially smaller than the number of episodes for executor training, the parts for hallucinator training are hardly noticeable.}
\label{fig:curve}
\end{figure*}

\begin{table}
\centering\small
\caption{Average rewards and average completion times on the test set for each approach at the end of training. We report the mean and standard error (in parentheses) over 5 random seeds for training.}
\label{tab:res}
\begin{tabular}{ccc|cc|cc}
\toprule
& \multicolumn{2}{c|}{\bf 2D-craft} & \multicolumn{2}{c}{\bf Box-world} & \multicolumn{2}{c}{\bf Ant-craft}\\
{} & {\bf Reward} & {\bf Finish step} & {\bf Reward} & {\bf Finish step} & {\bf Reward} & {\bf Finish step} \\
\midrule
End-to-end & 0.22 (0.01) & 82.3 (1.3) & 0.85 (0.02) & 44.7 (0.6) & 0.12 (0.03) & 93.1 (2.2) \\
World models \cite{world-models} & 0.23 (0.01) & 81.2 (0.7) & 0.80 (0.02) & 47.2 (0.9) & 0.13 (0.01) & 91.3 (1.2) \\
Relational \cite{drrl} & - & - & 0.77 (0.02) & 51.3 (1.6) & - & - \\
\midrule
Ours & 0.70 (0.03) & 56.4 (2.0) & 0.90 (0.00) & 38.6 (0.4) & 0.40 (0.01) & 79.2 (1.7) \\
Oracle & 0.76 (0.02) & 50.4 (1.1) & 0.97 (0.01) & 30.8 (0.5) & 0.43 (0.02) & 77.2 (1.6)\\
\bottomrule
\end{tabular}
\end{table}

\begin{figure}
\centering
\begin{subfigure}[]{0.05\textwidth}
\centering
\includegraphics[width=\textwidth]{figures/symbols-new.pdf}
\end{subfigure}
\begin{subfigure}[]{0.23\textwidth}
\centering
\includegraphics[width=\textwidth]{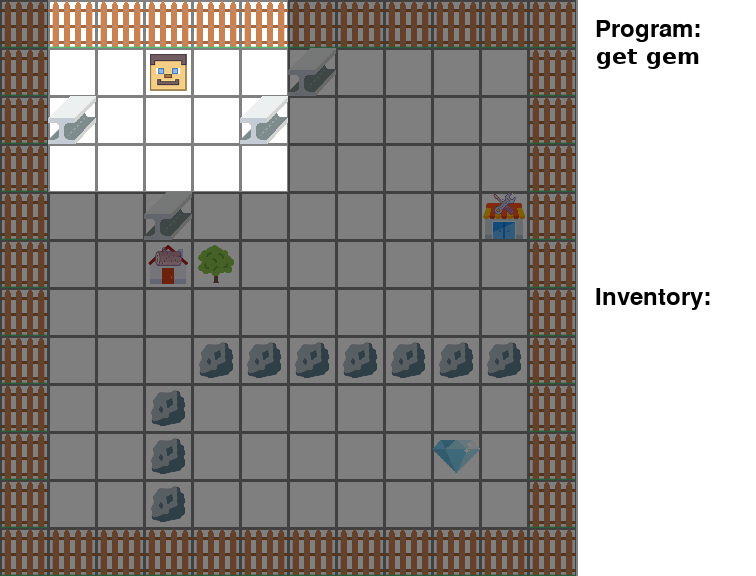}
\caption{}
\end{subfigure}
\begin{subfigure}[]{0.23\textwidth}
\centering
\includegraphics[width=\textwidth]{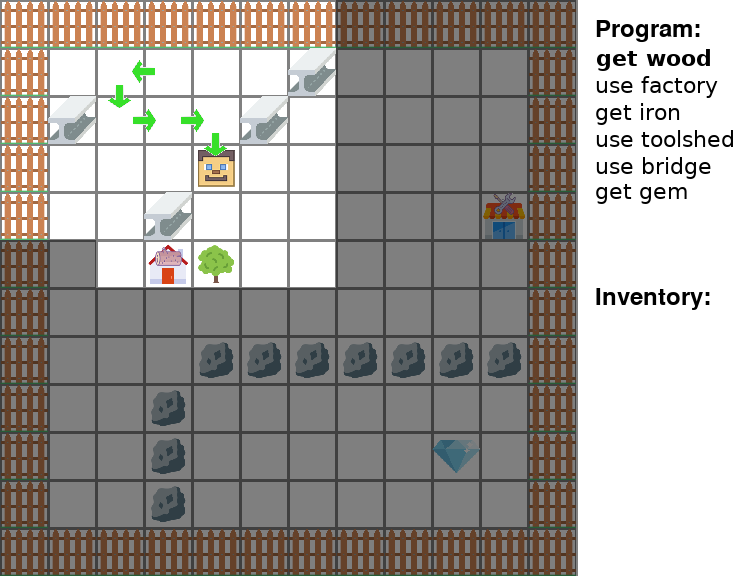}
\caption{}
\end{subfigure}
\begin{subfigure}[]{0.23\textwidth}
\centering
\includegraphics[width=\textwidth]{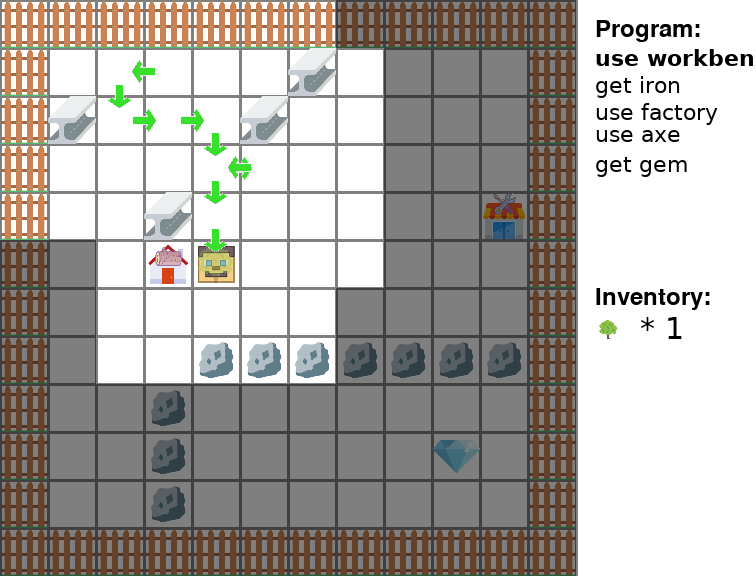}
\caption{}
\end{subfigure}
\begin{subfigure}[]{0.23\textwidth}
\centering
\includegraphics[width=\textwidth]{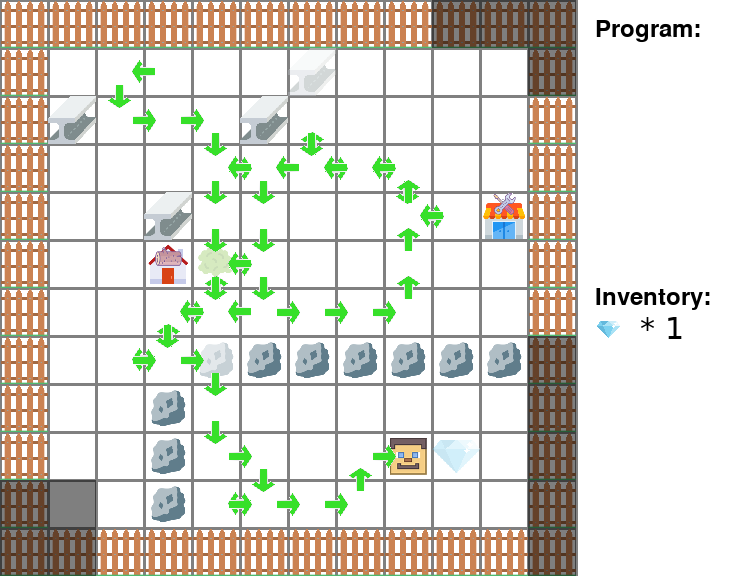}
\caption{}
\end{subfigure}
\caption{Example behavior of our policy in a task with the goal of getting gem.
(a) The start state. The agent initially hallucinates that there is a gem in the same zone, thus starts with a simple program ``get gem''. 
(b) After several steps, the agent observes a wood and a factory. Hallucinating based on these new observations, the agent synthesizes a new program that builds a bridge to cross some water and get gem. This is a reasonable guess since wood, iron and factory are part of the recipe to build a bridge, therefore the presence of them hints that the solution might be via building a bridge.
(c) After the agent finishes the ``get wood'' component, it observes that there are stones in the map, for which bridge cannot be used. Hallucinating based on these new observations, the agent synthesizes a new program that builds an axe to cross the stone. This is a correct program for this task. 
(d) The final state. The agent executes the program and successfully gets the gem.
}
\label{fig:example-traj}
\end{figure}

\subsection{Implementation Details}

\textbf{2D-craft.}
For our approach, we use a CVAE hallucinator, with MLP (with 200 hidden units) encoder/decoder, trained on 20K $(s, o)$ pairs collected by a random agent. We use the Z3~\cite{z3} solver to solve the MaxSAT problems. We use $m=3$ hallucinated environments, $N=20$ steps before replanning in our main experiments, and $N=5$ in the example behaviors we show for better demonstrations. We use the same actor (resp., critic) network architecture for the policies across all approaches---i.e., an MLP with 128 (resp., 32) hidden units.
We train the policies of each approach on 400K episodes over randomly sampled training tasks, and evaluate on a test set of 50 tasks.~\footnote{In our experiments, we train the hallucinator and the executor separately; but in general, one can also interleave the training of the two.}

\textbf{Box-world.}
Following \cite{drrl}, we use a one-layer CNN with 32 kernels of size $3\times3$ to preprocess the map across all approaches. For our approach, we have a component for each color where the subgoal is to get the key of that color; see Appendix \ref{append:box-semantics} for details. For the hallucinator, we use the same architecture as in the craft environment but with 300 hidden units, and trained with 100K $(s, o)$ pairs. For the synthesizer, we use $m=3$ and $N=10$. We train the policies for each approach on 200K episodes, and evaluate on a test set containing 40 tasks.

\subsection{Results}
\label{sec:results}

Table \ref{tab:res} (left two columns) shows the performance of each approach at the end of training. Figure \ref{fig:curve} shows the training curves. 
Our approach significantly outperforms the non-program-guided baselines, both in terms of fraction of tasks solved and in time taken to solve them; it also converges faster, demonstrating that program guidance makes learning significantly more tractable. Our approach also performs comparably to the oracle, delivering comparable performance with significantly less user burden. Figure \ref{fig:example-traj} shows the behavior of our policy in an example task in the 2D-craft environment; see Appendix \ref{append:additional-examples} for more examples.


\begin{figure*}
\centering
\begin{minipage}{0.45\linewidth}
\centering\small
\captionof{table}{Comparison to optimistic synthesis and random hallucination strategies on the 2D-craft environment.}
\label{tab:optim}
\begin{tabular}{ccc}
\toprule
{} & {\bf Avg. reward} & {\bf Avg. finish step} \\
\midrule
Ours & \textbf{0.70 (0.03)} & \textbf{56.4 (2.0)} \\
Optimistic & 0.42 (0.02) & 70.2 (1.2) \\
Random & 0.48 (0.02) & 72.6 (0.9) \\
\bottomrule
\end{tabular}
\end{minipage}
\qquad
\begin{minipage}{0.45\linewidth}
\centering
\begin{subfigure}[]{0.45\textwidth}
\centering
\includegraphics[width=\textwidth]{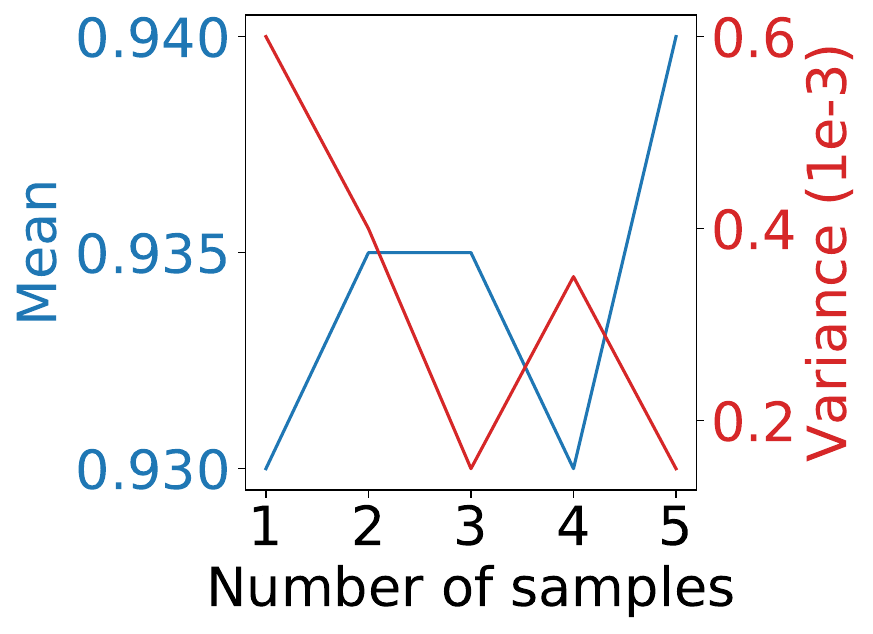}
\caption{Reward}
\end{subfigure}
\begin{subfigure}[]{0.45\textwidth}
\centering
\includegraphics[width=\textwidth]{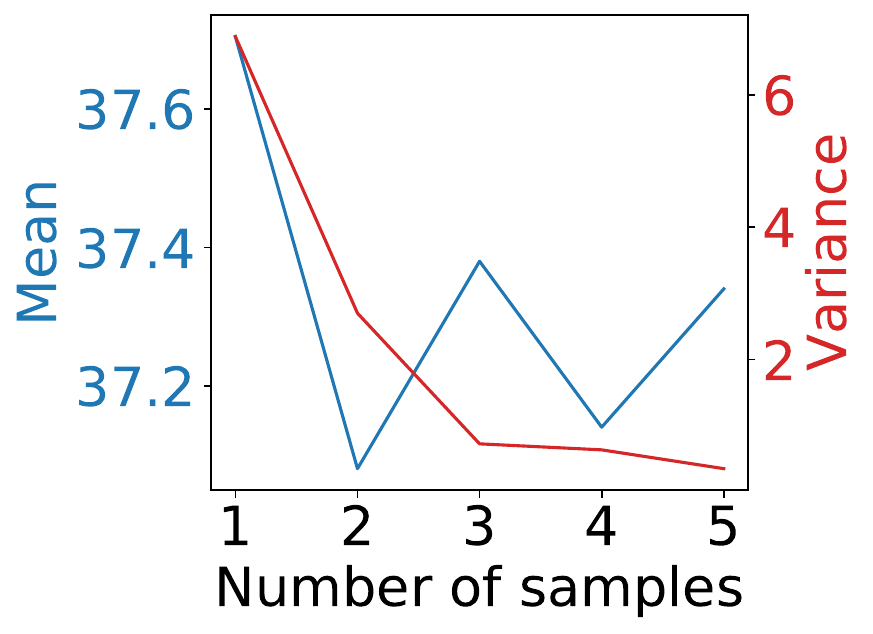}
\caption{Finish time}
\end{subfigure}
\caption{Effect of varying the number of samples $m$ on our approach, evaluated on the box-world over 5 random seeds. Mean and variance of (a) the average reward, (b) the average finishing time on the test tasks.}
\label{fig:n}
\end{minipage}
\end{figure*}

\textbf{Effect of the learned hallucinator.}
The hallucinator is a key in our approach to handle partial observations. Here we study the benefit of the learned hallucinator to our approach. 
First, we test a naive strategy for handling partial observations: the agent first randomly explores the map until the current zone is fully observed, then it synthesizes a program and follows it. This strategy only achieves an average reward of 0.024($\pm 0.004$) in 2D-craft, showing that our benchmarks require effective techniques for handling partial observations.
We compare to two ablations without a learned hallucinator: (i) an \emph{optimistic} synthesizer that synthesizes the shortest possible program making best-case assumptions about the unobserved parts of the map, and (ii) a \emph{random} hallucinator that randomly samples completions of the world (See Appendix \ref{append:abla} for more details). Table \ref{tab:optim} shows the results on the 2D-craft environment. As can be seen, our approach significantly outperforms both alternatives. Figure \ref{fig:ablation-optim} \& \ref{fig:ablation-ours} shows the difference in behavior between our approach and the optimistic strategy; by using a learned hallucinator, our approach is able to leverage the current observations effectively and synthesize a correct program sooner.

\textbf{Effect of the number of hallucinator samples.}
We vary the number of hallucinator samples $m$ on box-world. Figure \ref{fig:n} shows the results on the test set over 5 random seeds. As can be seen, varying $m$ does not significantly affect the mean performance, but increasing $m$ significantly reduces variance.
Thus, increasing $m$ makes the policy more robust to the uncertainty in the hallucinator. This fact shows the benefit of using multiple samples and MaxSAT synthesis.

\textbf{Transfer to MuJoCo Ant.}
To demonstrate that our approach can be adapted to handle continuous control tasks, we consider a variant of 2D-craft where the agent is replaced by a MuJoCo ant \cite{ant} (Figure~\ref{fig:ant-craft}).
We consider a simplified setup where we only model the movements of the ant; the ant automatically picks up resources in the grid cell it currently occupies.
We focus on transfer learning from 2D-craft. In particular, we pretrain a goal-reaching policy for the ant using soft actor-critic~\cite{sac}: given a random goal position, this policy moves the ant to that position. The actions output by each approach are translated into a goal position used as input to this goal-reaching policy. We initialize each policy with the corresponding model for 2D-craft and fine-tune it on ant-craft for 40K episodes. Table \ref{tab:res} (rightmost column) shows the results. Our approach significantly outperforms the non-program-guided baselines, both in terms of fraction of tasks solved and time taken to solve them. This demonstrates that our approach is also effective on tasks involving continuous control under a transfer learning setup. 
\section{Conclusion}
\label{sec:conclusions}

We propose an approach that automatically synthesizes programs to guide reinforcement learning for complex long-horizon tasks. Our model predictive program synthesis (MPPS) approach handles partially observed environments by leveraging an approach inspired by world models, where it learns a generative model over the remainder of the world conditioned on the observations, and then synthesizes a guiding program that accounts for the uncertainty in this model. Our experiments demonstrate that MPPS significantly outperforms non-program-guided approaches, while performing comparably to an oracle given a ground truth guiding program. Our results highlight that MPPS can deliver the benefits of program-guided reinforcement learning without requiring the user to provide a guiding program for every new task.

One limitation of our approach is that, as with existing program guided approaches, the user must provide a set of components for each domain. This process only needs to be completed once for each domain since the components can be reused across tasks; nevertheless, automatically inferring these components is an important direction for future work. Finally, we do not foresee any negative societal impacts or ethical concerns for our work (outside of generic risks in improving robotics capabilities).

\begin{ack}
We gratefully acknowledge support from DARPA HR001120C0015, NSF CCF-1917852, NSF CCF-1910769, and ARO W911NF-20-1-0080. The views expressed are those of the authors and do not reflect the official policy or position of the Department of Defense, the Army Research Office, or the U.S. Government. We thank the anonymous reviewers for their insightful and helpful comments.
\end{ack}

\bibliography{reference}

\begin{thebibliography}{79}
\providecommand{\natexlab}[1]{#1}
\providecommand{\url}[1]{\texttt{#1}}
\expandafter\ifx\csname urlstyle\endcsname\relax
  \providecommand{\doi}[1]{doi: #1}\else
  \providecommand{\doi}{doi: \begingroup \urlstyle{rm}\Url}\fi

\bibitem[Abel et~al.(2020)Abel, Umbanhowar, Khetarpal, Arumugam, Precup, and
  Littman]{abel2020value}
David Abel, Nate Umbanhowar, Khimya Khetarpal, Dilip Arumugam, Doina Precup,
  and Michael Littman.
\newblock Value preserving state-action abstractions.
\newblock In \emph{International Conference on Artificial Intelligence and
  Statistics}, pages 1639--1650. PMLR, 2020.

\bibitem[Anderson et~al.(2020)Anderson, Verma, Dillig, and
  Chaudhuri]{anderson2020neurosymbolic}
Greg Anderson, Abhinav Verma, Isil Dillig, and Swarat Chaudhuri.
\newblock Neurosymbolic reinforcement learning with formally verified
  exploration.
\newblock \emph{arXiv preprint arXiv:2009.12612}, 2020.

\bibitem[Andreas et~al.(2017)Andreas, Klein, and Levine]{psketch}
Jacob Andreas, Dan Klein, and Sergey Levine.
\newblock Modular multitask reinforcement learning with policy sketches.
\newblock In Doina Precup and Yee~Whye Teh, editors, \emph{Proceedings of the
  34th International Conference on Machine Learning}, volume~70 of
  \emph{Proceedings of Machine Learning Research}, pages 166--175,
  International Convention Centre, Sydney, Australia, 06--11 Aug 2017. PMLR.
\newblock URL \url{http://proceedings.mlr.press/v70/andreas17a.html}.

\bibitem[Arulkumaran et~al.(2017)Arulkumaran, Deisenroth, Brundage, and
  Bharath]{arulkumaran2017deep}
Kai Arulkumaran, Marc~Peter Deisenroth, Miles Brundage, and Anil~Anthony
  Bharath.
\newblock Deep reinforcement learning: A brief survey.
\newblock \emph{IEEE Signal Processing Magazine}, 34\penalty0 (6):\penalty0
  26--38, 2017.

\bibitem[Austin et~al.(2021)Austin, Odena, Nye, Bosma, Michalewski, Dohan,
  Jiang, Cai, Terry, Le, and Sutton]{largelm}
Jacob Austin, Augustus Odena, Maxwell Nye, Maarten Bosma, Henryk Michalewski,
  David Dohan, Ellen Jiang, Carrie~J. Cai, Michael Terry, Quoc~V. Le, and
  Charles Sutton.
\newblock Program synthesis with large language models.
\newblock \emph{CoRR}, abs/2108.07732, 2021.
\newblock URL \url{https://arxiv.org/abs/2108.07732}.

\bibitem[Bagaria and Konidaris(2020)]{Bagaria2020Option}
Akhil Bagaria and George Konidaris.
\newblock Option discovery using deep skill chaining.
\newblock In \emph{International Conference on Learning Representations}, 2020.
\newblock URL \url{https://openreview.net/forum?id=B1gqipNYwH}.

\bibitem[Balog et~al.(2017)Balog, Gaunt, Brockschmidt, Nowozin, and
  Tarlow]{deepcoder}
Matej Balog, Alexander~L. Gaunt, Marc Brockschmidt, Sebastian Nowozin, and
  Daniel Tarlow.
\newblock Deepcoder: Learning to write programs, 2017.

\bibitem[Bastani et~al.(2018)Bastani, Pu, and
  Solar-Lezama]{bastani2018verifiable}
Osbert Bastani, Yewen Pu, and Armando Solar-Lezama.
\newblock Verifiable reinforcement learning via policy extraction.
\newblock In \emph{Advances in neural information processing systems}, pages
  2494--2504, 2018.

\bibitem[Bonet(1998)]{bonet1998}
Blai Bonet.
\newblock High-level planning and control with incomplete information using
  pomdp's.
\newblock 1998.

\bibitem[Bunel et~al.(2018)Bunel, Hausknecht, Devlin, Singh, and
  Kohli]{bunel2018leveraging}
Rudy Bunel, Matthew Hausknecht, Jacob Devlin, Rishabh Singh, and Pushmeet
  Kohli.
\newblock Leveraging grammar and reinforcement learning for neural program
  synthesis.
\newblock In \emph{International Conference on Learning Representations}, 2018.
\newblock URL \url{https://openreview.net/forum?id=H1Xw62kRZ}.

\bibitem[Charlin et~al.(2007)Charlin, Poupart, and
  Shioda]{charlin2007automated}
Laurent Charlin, Pascal Poupart, and Romy Shioda.
\newblock Automated hierarchy discovery for planning in partially observable
  environments.
\newblock \emph{Advances in Neural Information Processing Systems},
  19:\penalty0 225, 2007.

\bibitem[Chen et~al.(2021)Chen, Lamoreaux, Wang, Durrett, Bastani, and
  Dillig]{chen2021web}
Qiaochu Chen, Aaron Lamoreaux, Xinyu Wang, Greg Durrett, Osbert Bastani, and
  Isil Dillig.
\newblock Web question answering with neurosymbolic program synthesis.
\newblock In \emph{Proceedings of the 42nd ACM SIGPLAN International Conference
  on Programming Language Design and Implementation}, pages 328--343, 2021.

\bibitem[Chen et~al.(2018)Chen, Liu, and Song]{chen2018synthesizing}
Xinyun Chen, Chang Liu, and Dawn Song.
\newblock Towards synthesizing complex programs from input-output examples,
  2018.

\bibitem[Chen et~al.(2019)Chen, Liu, and Song]{chen2018executionguided}
Xinyun Chen, Chang Liu, and Dawn Song.
\newblock Execution-guided neural program synthesis.
\newblock In \emph{International Conference on Learning Representations}, 2019.
\newblock URL \url{https://openreview.net/forum?id=H1gfOiAqYm}.

\bibitem[Chen et~al.(2020)Chen, Wang, Bastani, Dillig, and
  Feng]{chen2020program}
Yanju Chen, Chenglong Wang, Osbert Bastani, Isil Dillig, and Yu~Feng.
\newblock Program synthesis using deduction-guided reinforcement learning.
\newblock In \emph{International Conference on Computer Aided Verification},
  pages 587--610. Springer, 2020.

\bibitem[De~Moura and Bj\o{}rner(2008)]{z3}
Leonardo De~Moura and Nikolaj Bj\o{}rner.
\newblock Z3: An efficient smt solver.
\newblock In \emph{Proceedings of the Theory and Practice of Software, 14th
  International Conference on Tools and Algorithms for the Construction and
  Analysis of Systems}, TACAS'08/ETAPS'08, page 337–340, Berlin, Heidelberg,
  2008. Springer-Verlag.
\newblock ISBN 3540787992.

\bibitem[Devlin et~al.(2017)Devlin, Uesato, Bhupatiraju, Singh, Mohamed, and
  Kohli]{robustfill}
Jacob Devlin, Jonathan Uesato, Surya Bhupatiraju, Rishabh Singh, Abdel-rahman
  Mohamed, and Pushmeet Kohli.
\newblock Robustfill: Neural program learning under noisy i/o.
\newblock In \emph{Proceedings of the 34th International Conference on Machine
  Learning - Volume 70}, ICML'17, page 990–998. JMLR.org, 2017.

\bibitem[Draper et~al.(1994)Draper, Hanks, and Weld]{Draper1994}
Denise Draper, S.~Hanks, and Daniel~S. Weld.
\newblock Probabilistic planning with information gathering and contingent
  execution.
\newblock In \emph{AIPS}, 1994.

\bibitem[Ellis et~al.(2015)Ellis, Solar-Lezama, and
  Tenenbaum]{ellis2015unsupervised}
Kevin Ellis, Armando Solar-Lezama, and Josh Tenenbaum.
\newblock Unsupervised learning by program synthesis.
\newblock In \emph{Advances in neural information processing systems}, pages
  973--981, 2015.

\bibitem[Ellis et~al.(2018)Ellis, Ritchie, Solar-Lezama, and
  Tenenbaum]{ellis2018learning}
Kevin Ellis, Daniel Ritchie, Armando Solar-Lezama, and Josh Tenenbaum.
\newblock Learning to infer graphics programs from hand-drawn images.
\newblock In \emph{Advances in neural information processing systems}, pages
  6059--6068, 2018.

\bibitem[Ellis et~al.(2019)Ellis, Nye, Pu, Sosa, Tenenbaum, and
  Solar-Lezama]{ellis2019write}
Kevin Ellis, Maxwell Nye, Yewen Pu, Felix Sosa, Josh Tenenbaum, and Armando
  Solar-Lezama.
\newblock Write, execute, assess: Program synthesis with a repl.
\newblock In \emph{Advances in Neural Information Processing Systems}, pages
  9169--9178, 2019.

\bibitem[Ellis et~al.(2021)Ellis, Wong, Nye, Sabl{\'e}-Meyer, Morales, Hewitt,
  Cary, Solar-Lezama, and Tenenbaum]{ellis2021dreamcoder}
Kevin Ellis, Catherine Wong, Maxwell Nye, Mathias Sabl{\'e}-Meyer, Lucas
  Morales, Luke Hewitt, Luc Cary, Armando Solar-Lezama, and Joshua~B Tenenbaum.
\newblock Dreamcoder: bootstrapping inductive program synthesis with wake-sleep
  library learning.
\newblock In \emph{Proceedings of the 42nd ACM SIGPLAN International Conference
  on Programming Language Design and Implementation}, pages 835--850, 2021.

\bibitem[Feng et~al.(2018)Feng, Martins, Bastani, and Dillig]{feng2018program}
Yu~Feng, Ruben Martins, Osbert Bastani, and Isil Dillig.
\newblock Program synthesis using conflict-driven learning.
\newblock \emph{ACM SIGPLAN Notices}, 53\penalty0 (4):\penalty0 420--435, 2018.

\bibitem[Fikes and Nilsson(1971)]{fikes1971strips}
Richard~E Fikes and Nils~J Nilsson.
\newblock Strips: A new approach to the application of theorem proving to
  problem solving.
\newblock \emph{Artificial intelligence}, 2\penalty0 (3-4):\penalty0 189--208,
  1971.

\bibitem[Finn et~al.(2017)Finn, Abbeel, and Levine]{finn2017model}
Chelsea Finn, Pieter Abbeel, and Sergey Levine.
\newblock Model-agnostic meta-learning for fast adaptation of deep networks.
\newblock In \emph{International Conference on Machine Learning}, pages
  1126--1135. PMLR, 2017.

\bibitem[Garrett et~al.(2020)Garrett, Chitnis, Holladay, Kim, Silver,
  Kaelbling, and Lozano{-}P{\'{e}}rez]{inttamp}
Caelan~Reed Garrett, Rohan Chitnis, Rachel Holladay, Beomjoon Kim, Tom Silver,
  Leslie~Pack Kaelbling, and Tom{\'{a}}s Lozano{-}P{\'{e}}rez.
\newblock Integrated task and motion planning.
\newblock \emph{CoRR}, abs/2010.01083, 2020.
\newblock URL \url{https://arxiv.org/abs/2010.01083}.

\bibitem[Groshev et~al.(2018)Groshev, Goldstein, Tamar, Srivastava, and
  Abbeel]{genplan18}
Edward Groshev, Maxwell Goldstein, Aviv Tamar, Siddharth Srivastava, and Pieter
  Abbeel.
\newblock Learning generalized reactive policies using deep neural networks,
  2018.

\bibitem[Gulwani(2011)]{flashfill}
Sumit Gulwani.
\newblock Automating string processing in spreadsheets using input-output
  examples.
\newblock In \emph{PoPL'11, January 26-28, 2011, Austin, Texas, USA}, January
  2011.

\bibitem[Ha and Schmidhuber(2018)]{world-models}
David Ha and J{\"{u}}rgen Schmidhuber.
\newblock World models.
\newblock \emph{CoRR}, abs/1803.10122, 2018.
\newblock URL \url{http://arxiv.org/abs/1803.10122}.

\bibitem[Haarnoja et~al.(2018)Haarnoja, Zhou, Abbeel, and Levine]{sac}
Tuomas Haarnoja, Aurick Zhou, Pieter Abbeel, and Sergey Levine.
\newblock Soft actor-critic: Off-policy maximum entropy deep reinforcement
  learning with a stochastic actor.
\newblock In Jennifer Dy and Andreas Krause, editors, \emph{Proceedings of the
  35th International Conference on Machine Learning}, volume~80 of
  \emph{Proceedings of Machine Learning Research}, pages 1861--1870,
  Stockholmsmässan, Stockholm Sweden, 10--15 Jul 2018. PMLR.
\newblock URL \url{http://proceedings.mlr.press/v80/haarnoja18b.html}.

\bibitem[Handa and Rinard(2020)]{shivam}
Shivam Handa and Martin Rinard.
\newblock Inductive program synthesis over noisy data.
\newblock \emph{CoRR}, abs/2009.10272, 2020.
\newblock URL \url{https://arxiv.org/abs/2009.10272}.

\bibitem[Hasanbeig et~al.(2019)Hasanbeig, Jeppu, Abate, Melham, and
  Kroening]{deepsynth}
Mohammadhosein Hasanbeig, Natasha~Yogananda Jeppu, Alessandro Abate, Tom
  Melham, and Daniel Kroening.
\newblock Deepsynth: Program synthesis for automatic task segmentation in deep
  reinforcement learning.
\newblock \emph{CoRR}, abs/1911.10244, 2019.
\newblock URL \url{http://arxiv.org/abs/1911.10244}.

\bibitem[Hausman et~al.(2018)Hausman, Springenberg, Wang, Heess, and
  Riedmiller]{hausman2018learning}
Karol Hausman, Jost~Tobias Springenberg, Ziyu Wang, Nicolas Heess, and Martin
  Riedmiller.
\newblock Learning an embedding space for transferable robot skills.
\newblock In \emph{International Conference on Learning Representations}, 2018.

\bibitem[Hu and De~Giacomo(2011)]{genplan11}
Yuxiao Hu and Giuseppe De~Giacomo.
\newblock Generalized planning: Synthesizing plans that work for multiple
  environments.
\newblock In \emph{Proceedings of the Twenty-Second International Joint
  Conference on Artificial Intelligence - Volume Volume Two}, IJCAI'11, page
  918–923. AAAI Press, 2011.
\newblock ISBN 9781577355144.

\bibitem[Huang et~al.(2020)Huang, Smith, Bastani, Singh, Albarghouthi, and
  Naik]{huang2020generating}
Jiani Huang, Calvin Smith, Osbert Bastani, Rishabh Singh, Aws Albarghouthi, and
  Mayur Naik.
\newblock Generating programmatic referring expressions via program synthesis.
\newblock In \emph{International Conference on Machine Learning}, pages
  4495--4506. PMLR, 2020.

\bibitem[Illanes and McIlraith(2019)]{genplan19}
León Illanes and Sheila~A. McIlraith.
\newblock Generalized planning via abstraction: Arbitrary numbers of objects.
\newblock \emph{Proceedings of the AAAI Conference on Artificial Intelligence},
  33\penalty0 (01):\penalty0 7610--7618, Jul. 2019.
\newblock \doi{10.1609/aaai.v33i01.33017610}.
\newblock URL \url{https://ojs.aaai.org/index.php/AAAI/article/view/4754}.

\bibitem[Inala et~al.(2020{\natexlab{a}})Inala, Bastani, Tavares, and
  Solar-Lezama]{inala2020synthesizing}
Jeevana~Priya Inala, Osbert Bastani, Zenna Tavares, and Armando Solar-Lezama.
\newblock Synthesizing programmatic policies that inductively generalize.
\newblock In \emph{International Conference on Learning Representations},
  2020{\natexlab{a}}.

\bibitem[Inala et~al.(2020{\natexlab{b}})Inala, Yang, Paulos, Pu, Bastani,
  Kumar, Rinard, and Solar-Lezama]{inala2021neurosymbolic}
Jeevana~Priya Inala, Yichen Yang, James Paulos, Yewen Pu, Osbert Bastani, Vijay
  Kumar, Martin Rinard, and Armando Solar-Lezama.
\newblock Neurosymbolic transformers for multi-agent communication.
\newblock In \emph{NeurIPS}, 2020{\natexlab{b}}.

\bibitem[Jothimurugan et~al.(2019)Jothimurugan, Alur, and
  Bastani]{jothimurugan2019composable}
Kishor Jothimurugan, Rajeev Alur, and Osbert Bastani.
\newblock A composable specification language for reinforcement learning tasks.
\newblock In \emph{NeurIPS}, 2019.

\bibitem[Jothimurugan et~al.(2021{\natexlab{a}})Jothimurugan, Bansal, Bastani,
  and Alur]{jothimurugan2021compositional}
Kishor Jothimurugan, Suguman Bansal, Osbert Bastani, and Rajeev Alur.
\newblock Compositional reinforcement learning from logical specifications.
\newblock \emph{arXiv preprint arXiv:2106.13906}, 2021{\natexlab{a}}.

\bibitem[Jothimurugan et~al.(2021{\natexlab{b}})Jothimurugan, Bastani, and
  Alur]{jothimurugan2021abstract}
Kishor Jothimurugan, Osbert Bastani, and Rajeev Alur.
\newblock Abstract value iteration for hierarchical reinforcement learning.
\newblock In \emph{AISTATS}, 2021{\natexlab{b}}.

\bibitem[Kaelbling and Lozano-Pérez(2011)]{tampnow}
Leslie~Pack Kaelbling and Tomás Lozano-Pérez.
\newblock Hierarchical task and motion planning in the now.
\newblock In \emph{2011 IEEE International Conference on Robotics and
  Automation}, pages 1470--1477, 2011.
\newblock \doi{10.1109/ICRA.2011.5980391}.

\bibitem[Kaelbling and Lozano-Pérez(2013)]{tampbelief}
Leslie~Pack Kaelbling and Tomás Lozano-Pérez.
\newblock Integrated task and motion planning in belief space.
\newblock \emph{The International Journal of Robotics Research}, 32\penalty0
  (9-10):\penalty0 1194--1227, 2013.
\newblock \doi{10.1177/0278364913484072}.

\bibitem[Kalyan et~al.(2018)Kalyan, Mohta, Polozov, Batra, Jain, and
  Gulwani]{kalyan2018neuralguided}
Ashwin Kalyan, Abhishek Mohta, Oleksandr Polozov, Dhruv Batra, Prateek Jain,
  and Sumit Gulwani.
\newblock Neural-guided deductive search for real-time program synthesis from
  examples.
\newblock In \emph{International Conference on Learning Representations}, 2018.
\newblock URL \url{https://openreview.net/forum?id=rywDjg-RW}.

\bibitem[Kingma and Ba(2017)]{kingma2017adam}
Diederik~P. Kingma and Jimmy Ba.
\newblock Adam: A method for stochastic optimization, 2017.

\bibitem[Kingma and Welling(2013)]{kingma2013auto}
Diederik~P Kingma and Max Welling.
\newblock Auto-encoding variational bayes.
\newblock \emph{arXiv preprint arXiv:1312.6114}, 2013.

\bibitem[Konda and Tsitsiklis(2000)]{ac}
Vijay Konda and John Tsitsiklis.
\newblock Actor-critic algorithms.
\newblock In S.~Solla, T.~Leen, and K.~M\"{u}ller, editors, \emph{Advances in
  Neural Information Processing Systems}, volume~12, pages 1008--1014. MIT
  Press, 2000.
\newblock URL
  \url{https://proceedings.neurips.cc/paper/1999/file/6449f44a102fde848669bdd9eb6b76fa-Paper.pdf}.

\bibitem[Krentel(1986)]{maxsat}
M~W Krentel.
\newblock The complexity of optimization problems.
\newblock In \emph{Proceedings of the Eighteenth Annual ACM Symposium on Theory
  of Computing}, STOC '86, page 69–76, New York, NY, USA, 1986. Association
  for Computing Machinery.
\newblock ISBN 0897911938.
\newblock \doi{10.1145/12130.12138}.
\newblock URL \url{https://doi.org/10.1145/12130.12138}.

\bibitem[Li et~al.(2020)Li, Florensa, Clavera, and Abbeel]{li2020subpolicy}
Alexander~C. Li, Carlos Florensa, Ignasi Clavera, and Pieter Abbeel.
\newblock Sub-policy adaptation for hierarchical reinforcement learning, 2020.

\bibitem[Mnih et~al.(2015)Mnih, Kavukcuoglu, Silver, Rusu, Veness, Bellemare,
  Graves, Riedmiller, Fidjeland, Ostrovski, et~al.]{mnih2015human}
Volodymyr Mnih, Koray Kavukcuoglu, David Silver, Andrei~A Rusu, Joel Veness,
  Marc~G Bellemare, Alex Graves, Martin Riedmiller, Andreas~K Fidjeland, Georg
  Ostrovski, et~al.
\newblock Human-level control through deep reinforcement learning.
\newblock \emph{nature}, 518\penalty0 (7540):\penalty0 529--533, 2015.

\bibitem[Nye et~al.(2020)Nye, Pu, Bowers, Andreas, Tenenbaum, and
  Solar-Lezama]{nye2020representing}
Maxwell Nye, Yewen Pu, Matthew Bowers, Jacob Andreas, Joshua~B Tenenbaum, and
  Armando Solar-Lezama.
\newblock Representing partial programs with blended abstract semantics.
\newblock \emph{arXiv preprint arXiv:2012.12964}, 2020.

\bibitem[Phiquepal and Toussaint(2019)]{tamppart}
Camille Phiquepal and Marc Toussaint.
\newblock Combined task and motion planning under partial observability: An
  optimization-based approach.
\newblock In \emph{2019 International Conference on Robotics and Automation
  (ICRA)}, pages 9000--9006, 2019.
\newblock \doi{10.1109/ICRA.2019.8793260}.

\bibitem[Schulman et~al.(2016)Schulman, Moritz, Levine, Jordan, and
  Abbeel]{ant}
John Schulman, Philipp Moritz, Sergey Levine, Michael Jordan, and Pieter
  Abbeel.
\newblock High-dimensional continuous control using generalized advantage
  estimation.
\newblock In \emph{Proceedings of the International Conference on Learning
  Representations (ICLR)}, 2016.

\bibitem[Shah et~al.(2020)Shah, Zhan, Sun, Verma, Yue, and
  Chaudhuri]{admissible}
Ameesh Shah, Eric Zhan, Jennifer~J. Sun, Abhinav Verma, Yisong Yue, and Swarat
  Chaudhuri.
\newblock Learning differentiable programs with admissible neural heuristics.
\newblock \emph{CoRR}, abs/2007.12101, 2020.
\newblock URL \url{https://arxiv.org/abs/2007.12101}.

\bibitem[Shaw et~al.(1975)Shaw, Swartout, and Green]{deshaw}
David~E. Shaw, William~R. Swartout, and C.~Cordell Green.
\newblock Inferring lisp programs from examples.
\newblock In \emph{Proceedings of the 4th International Joint Conference on
  Artificial Intelligence - Volume 1}, IJCAI'75, page 260–267, San Francisco,
  CA, USA, 1975. Morgan Kaufmann Publishers Inc.

\bibitem[Sohn et~al.(2015)Sohn, Lee, and Yan]{cvae}
Kihyuk Sohn, Honglak Lee, and Xinchen Yan.
\newblock Learning structured output representation using deep conditional
  generative models.
\newblock In C.~Cortes, N.~Lawrence, D.~Lee, M.~Sugiyama, and R.~Garnett,
  editors, \emph{Advances in Neural Information Processing Systems}, volume~28,
  pages 3483--3491. Curran Associates, Inc., 2015.
\newblock URL
  \url{https://proceedings.neurips.cc/paper/2015/file/8d55a249e6baa5c06772297520da2051-Paper.pdf}.

\bibitem[Sohn et~al.(2018)Sohn, Oh, and Lee]{Sohn18}
Sungryull Sohn, Junhyuk Oh, and Honglak Lee.
\newblock Hierarchical reinforcement learning for zero-shot generalization with
  subtask dependencies.
\newblock In \emph{Proceedings of the 32nd International Conference on Neural
  Information Processing Systems}, NIPS'18, page 7156–7166, Red Hook, NY,
  USA, 2018. Curran Associates Inc.

\bibitem[Solar-Lezama(2008)]{sketch}
Armando Solar-Lezama.
\newblock \emph{Program Synthesis by Sketching}.
\newblock PhD thesis, USA, 2008.

\bibitem[Srivastava(2011)]{genplanthesis}
Siddharth Srivastava.
\newblock Foundations and applications of generalized planning.
\newblock \emph{AI Commun.}, 24\penalty0 (4):\penalty0 349–351, December
  2011.
\newblock ISSN 0921-7126.

\bibitem[Stentz et~al.(1995)]{stentz1995focussed}
Anthony Stentz et~al.
\newblock The focussed d\^{}* algorithm for real-time replanning.
\newblock In \emph{IJCAI}, volume~95, pages 1652--1659, 1995.

\bibitem[Stolle and Precup(2002)]{stolle2002learning}
Martin Stolle and Doina Precup.
\newblock Learning options in reinforcement learning.
\newblock In \emph{International Symposium on abstraction, reformulation, and
  approximation}, pages 212--223. Springer, 2002.

\bibitem[Sun et~al.(2020)Sun, Wu, and Lim]{pagent}
Shao-Hua Sun, Te-Lin Wu, and Joseph~J. Lim.
\newblock Program guided agent.
\newblock In \emph{International Conference on Learning Representations}, 2020.
\newblock URL \url{https://openreview.net/forum?id=BkxUvnEYDH}.

\bibitem[Sutton et~al.(1999)Sutton, Precup, and Singh]{sutton1999between}
Richard~S Sutton, Doina Precup, and Satinder Singh.
\newblock Between mdps and semi-mdps: A framework for temporal abstraction in
  reinforcement learning.
\newblock \emph{Artificial intelligence}, 112\penalty0 (1-2):\penalty0
  181--211, 1999.

\bibitem[Tessler et~al.(2016)Tessler, Givony, Zahavy, Mankowitz, and
  Mannor]{tessler2016deep}
Chen Tessler, Shahar Givony, Tom Zahavy, Daniel~J. Mankowitz, and Shie Mannor.
\newblock A deep hierarchical approach to lifelong learning in minecraft, 2016.

\bibitem[Tian et~al.(2019)Tian, Luo, Sun, Ellis, Freeman, Tenenbaum, and
  Wu]{tian2019learning}
Yonglong Tian, Andrew Luo, Xingyuan Sun, Kevin Ellis, William~T Freeman,
  Joshua~B Tenenbaum, and Jiajun Wu.
\newblock Learning to infer and execute 3d shape programs.
\newblock \emph{arXiv preprint arXiv:1901.02875}, 2019.

\bibitem[{Todorov} et~al.(2012){Todorov}, {Erez}, and {Tassa}]{mujoco}
E.~{Todorov}, T.~{Erez}, and Y.~{Tassa}.
\newblock Mujoco: A physics engine for model-based control.
\newblock In \emph{2012 IEEE/RSJ International Conference on Intelligent Robots
  and Systems}, pages 5026--5033, 2012.
\newblock \doi{10.1109/IROS.2012.6386109}.

\bibitem[Toussaint et~al.(2008)Toussaint, Charlin, and
  Poupart]{toussaint2008hierarchical}
Marc Toussaint, Laurent Charlin, and Pascal Poupart.
\newblock Hierarchical pomdp controller optimization by likelihood
  maximization.
\newblock In \emph{UAI}, volume~24, pages 562--570, 2008.

\bibitem[Valkov et~al.(2018)Valkov, Chaudhari, Srivastava, Sutton, and
  Chaudhuri]{valkov2018houdini}
Lazar Valkov, Dipak Chaudhari, Akash Srivastava, Charles Sutton, and Swarat
  Chaudhuri.
\newblock Houdini: Lifelong learning as program synthesis.
\newblock In \emph{Advances in Neural Information Processing Systems}, pages
  8687--8698, 2018.

\bibitem[Vaswani et~al.(2017)Vaswani, Shazeer, Parmar, Uszkoreit, Jones, Gomez,
  Kaiser, and Polosukhin]{transformer}
Ashish Vaswani, Noam Shazeer, Niki Parmar, Jakob Uszkoreit, Llion Jones,
  Aidan~N. Gomez, Lukasz Kaiser, and Illia Polosukhin.
\newblock Attention is all you need, 2017.

\bibitem[Verma(2019)]{verma2019verifiable}
Abhinav Verma.
\newblock Verifiable and interpretable reinforcement learning through program
  synthesis.
\newblock In \emph{Proceedings of the AAAI Conference on Artificial
  Intelligence}, volume~33, pages 9902--9903, 2019.

\bibitem[Verma et~al.(2018)Verma, Murali, Singh, Kohli, and
  Chaudhuri]{verma2018programmatically}
Abhinav Verma, Vijayaraghavan Murali, Rishabh Singh, Pushmeet Kohli, and Swarat
  Chaudhuri.
\newblock Programmatically interpretable reinforcement learning.
\newblock In \emph{International Conference on Machine Learning}, pages
  5045--5054. PMLR, 2018.

\bibitem[Verma et~al.(2019)Verma, Le, Yue, and Chaudhuri]{verma2019imitation}
Abhinav Verma, Hoang~M Le, Yisong Yue, and Swarat Chaudhuri.
\newblock Imitation-projected programmatic reinforcement learning.
\newblock In \emph{NeurIPS}, 2019.

\bibitem[Wang et~al.(2017)Wang, Dillig, and Singh]{treeautomata}
Xinyu Wang, Isil Dillig, and Rishabh Singh.
\newblock Synthesis of data completion scripts using finite tree automata.
\newblock \emph{Proc. ACM Program. Lang.}, 1\penalty0 (OOPSLA), October 2017.

\bibitem[Wulfmeier et~al.(2021)Wulfmeier, Rao, Hafner, Lampe, Abdolmaleki,
  Hertweck, Neunert, Tirumala, Siegel, Heess, and Riedmiller]{wulfmeier2021}
Markus Wulfmeier, Dushyant Rao, Roland Hafner, Thomas Lampe, Abbas Abdolmaleki,
  Tim Hertweck, Michael Neunert, Dhruva Tirumala, Noah Siegel, Nicolas Heess,
  and Martin Riedmiller.
\newblock Data-efficient hindsight off-policy option learning, 2021.

\bibitem[Young et~al.(2019)Young, Bastani, and Naik]{young2019learning}
Halley Young, Osbert Bastani, and Mayur Naik.
\newblock Learning neurosymbolic generative models via program synthesis.
\newblock In \emph{ICML}, 2019.

\bibitem[Zambaldi et~al.(2019)Zambaldi, Raposo, Santoro, Bapst, Li, Babuschkin,
  Tuyls, Reichert, Lillicrap, Lockhart, Shanahan, Langston, Pascanu, Botvinick,
  Vinyals, and Battaglia]{drrl}
Vinicius Zambaldi, David Raposo, Adam Santoro, Victor Bapst, Yujia Li, Igor
  Babuschkin, Karl Tuyls, David Reichert, Timothy Lillicrap, Edward Lockhart,
  Murray Shanahan, Victoria Langston, Razvan Pascanu, Matthew Botvinick, Oriol
  Vinyals, and Peter Battaglia.
\newblock Deep reinforcement learning with relational inductive biases.
\newblock In \emph{International Conference on Learning Representations}, 2019.
\newblock URL \url{https://openreview.net/forum?id=HkxaFoC9KQ}.

\bibitem[Zhang et~al.(2021)Zhang, Yu, and Xu]{zhang2021hier}
Jesse Zhang, Haonan Yu, and Wei Xu.
\newblock Hierarchical reinforcement learning by discovering intrinsic options,
  2021.

\bibitem[Zhang et~al.(2018)Zhang, Rosenblatt, Fetaya, Liao, Byrd, Might,
  Urtasun, and Zemel]{neuralguided}
Lisa Zhang, Gregory Rosenblatt, Ethan Fetaya, Renjie Liao, William~E. Byrd,
  Matthew Might, Raquel Urtasun, and Richard~S. Zemel.
\newblock Neural guided constraint logic programming for program synthesis.
\newblock \emph{CoRR}, abs/1809.02840, 2018.
\newblock URL \url{http://arxiv.org/abs/1809.02840}.

\bibitem[Zhang and Whiteson(2019)]{dac2019}
Shangtong Zhang and Shimon Whiteson.
\newblock Dac: The double actor-critic architecture for learning options, 2019.

\end{thebibliography}
\bibliographystyle{plainnat}

\clearpage
\appendix
\section{Components for Environments}

\subsection{Components for the Craft Environment}
\label{append:semantics}

In this section, we describe the components (i.e., logical formulae encoding pre/post-conditions for each option) that we use for the craft environment. First, recall that the domain-specific language that encodes the set of components for the craft environment is
\[
\small
\begin{array}{rcl}
C &:=& \textrm{get} ~R ~\vert~ \textrm{use} ~T ~\vert~ \textrm{use} ~W \\
R &:=& \textrm{wood} ~\vert~ \textrm{iron} ~\vert~ \textrm{grass} ~\vert~ \textrm{gold} ~\vert~ \textrm{gem} \\
T &:=& \textrm{bridge} ~\vert~ \textrm{axe} ~\vert~ \textrm{ladder} \\
W &:=& \textrm{factory} ~\vert~ \textrm{workbench} ~\vert~ \textrm{toolshed} \\
\end{array}
\]
Also, the set of possible artifacts (objects that can be made in some workshop using resources or other artifacts) in the craft environment is
\begin{align*}
A = \left\{\begin{array}{c}
\text{bridge},~\text{axe},~\text{plank},~\text{stick},~\text{ladder}
\end{array}\right\}. 
\end{align*}
We define the following features:
\begin{itemize}[topsep=0pt,itemsep=0ex,partopsep=1ex,parsep=1ex]
\item \textbf{Zone:} $z=i$ indicates the agent is in zone $i$
\item \textbf{Boundary:} $b_{i,j}=b$ indicates how zones $i$ and $j$ are connected, where
\begin{align*}
b\in\{\text{connected},\text{water},\text{stone},\text{not adjacent}\}
\end{align*}
\item \textbf{Resource:} $\rho_{i,r}=n$ indicates that there are $n$ units of resource $r$ in zone $i$
\item \textbf{Workshop:} $\omega_{i,r}=b$, where $b\in\{\text{true},\text{false}\}$, indicates whether there exists a workshop $r$ in zone $i$
\item \textbf{Inventory:} $\iota_r=n$ indicates that there are $n$ objects $r$ (either a resource or an artifact) in the agent's inventory
\end{itemize}
We use $z^-, b^-, \rho^-, \omega^-, \iota^-$ and $z^+, b^+, \rho^+, \omega^+, \iota^+$ to denote the initial state and the final state for a component, respectively. Now, the logical formulae for each component are defined as follows.

\textbf{(1) ``get $r$'' (for any resource $r\in R$).}
First, we have the following component telling the agent to obtain a specific resource $r$:
\begin{align*}
\forall i, j\;.\; & ( z^- = i \wedge z^+ = j ) \Rightarrow ( b_{i,j}^- = \text{connected} ) \\
&~\wedge ( \rho^+_{j,r} = \rho_{j,r}^- - 1 ) \wedge ( \iota^+_{r} = \iota_{r}^- + 1 ) \wedge \mathcal{Q}.
\end{align*}
Here, $\mathcal{Q}$ refers to the conditions that the other fields of the abstract state stay the same---i.e.,
\begin{align*}
&(b^+ = b^-) \wedge (\omega^+ = \omega^-) \wedge (\iota_{\setminus r}^+ = \iota_{\setminus r}^-) \\
& \qquad \wedge (\rho_{\setminus (j,r)}^+ = \rho_{\setminus (j,r)}^-),
\end{align*}
where $\iota_{\setminus r}$ means all the other fields in $\iota$ except $\iota_{r}$, and similarly for $\rho_{\setminus (j,r)}$. In particular $\mathcal{Q}$ addresses the \emph{frame problem} from classical planning.

\textbf{(2) ``use $r$'' (for any workshop $r \in W$).}
Next, we have a component telling the agent to use a workshop to create an artifact. To do so, we introduce a set of auxiliary features to denote the number of artifacts made in this component: $m_o = n$ indicates that $n$ units of artifact $o$ is made. The set of artifacts that can be made at workshop $r$ is denoted as $A_r$, and the number of units of ingredient $q$ needed to make 1 unit of artifact $o$ is denoted as $k_{o, q}$, where $q\in R \cup A$; note that $\{A_r\}$ and $\{k_{o, q}\}$ come from the rule of the game. 

Then, the logical formula for ``use $r$'' is
\begin{align*}
\forall i &, j\;.\; ( z^- = i \wedge z^+ = j ) \Rightarrow ( b_{i,j}^- = \text{connected} ) \\
&~\wedge ( w_{j, r} = \text{true} ) \wedge \left( \sum_{o\in A_r} m_o \geq 1 \right) \wedge \left( \sum_{o\notin A_r} m_o = 0 \right) \\
&~\wedge \left( \forall q \in R, ~\iota_q^+ = \iota_q^- - \sum_{o \in A_r} k_{o,q} m_o \right) \\
&~\wedge \left( \forall q \in A, ~\iota_q^+ = \iota_q^- - \sum_{o \in A_r} k_{o,q} m_o + m_q \right) \\
&~\wedge \left( \forall o \in A_r, ~\lnot\left( \bigwedge_{q} \iota_q^+ \geq k_{o,q} \right) \right) \\
&~\wedge \mathcal{Q},
\end{align*}
where
\begin{align*}
\mathcal{Q} &= (b^+ = b^-) \wedge (\omega^+ = \omega^-) \wedge (\rho^+ = \rho^-).
\end{align*}
This formula reflects the game setting that when the agent uses a workshop, it will make artifacts until the ingredients in the inventory are depleted. 

\textbf{(3) ``use r'' ($r=$ bridge/axe/ladder).}
Next, we have the following component for telling the agent to use a tool. The formula for this component encodes the logic of zone connectivity. In particular, it is
\begin{align*}
\forall i &, j\;.\; ( z^- = i \wedge z^+ = j ) \Rightarrow ( b_{i,j}^- = \text{water/stone} ) \\
&~\wedge ( b_{i,j}^+ = \text{connected} ) 
\wedge ( \iota_r^+ = \iota_r^- - 1 ) \\
&~\wedge \Big( \forall i', j', ~(b_{i',j'}^+ = \text{connected}) \Rightarrow \\
&\qquad\qquad\qquad \big( ( b_{i',j'}^- = \text{connected} ) \lor \mathcal{X} \big) \Big) \\
&~\wedge \Big( \forall i', j', ~(b_{i',j'}^+ \neq \text{connected}) \Rightarrow ( b_{i',j'}^+ = b_{i',j'}^- ) \Big) \\
&~\wedge \mathcal{Q},
\end{align*}
where
\begin{align*}
\mathcal{X} &= ( b_{i', i}^- = \text{connected} \lor  b_{i', j}^- = \text{connected} )\\
&~~~\wedge ( b_{j', i}^- = \text{connected} \lor  b_{j', j}^- = \text{connected} ) \\
\mathcal{Q} &= (\omega^+ = \omega^-) \wedge (\rho^+ = \rho^-) \wedge ( \iota_{\setminus r}^+ = \iota_{\setminus r}^- ).
\end{align*}

\subsection{Components for Box World}
\label{append:box-semantics}

In this section, we describe the components for the box world. They are all of the form ``get $k$'', where $k \in K$ is a color in the set of possible colors in the box world. First, we define the following features:
\begin{itemize}[topsep=0pt,itemsep=0ex,partopsep=1ex,parsep=1ex]
\item \textbf{Box}: $b_{k_1, k_2} = n$ indicates that there are $n$ boxes with key color $k_1$ and lock color $k_2$ in the map
\item \textbf{Loose key}: $\ell_k = b$, where $b\in \{\text{true}, \text{false}\}$, indicates whether there exists a loose key of color $k$ in the map
\item \textbf{Agent's key}: $\iota_k = b$, where $b\in \{\text{true}, \text{false}\}$, indicates whether the agent holds a key of color $k$
\end{itemize}

As in the craft environment, we use $ b^-, \ell^-, \iota^-$ and $ b^+, \ell^+, \iota^+$ to denote the initial state and the final state for a component, respectively. Since the configurations of the map in the box world can only contain at most one loose key, we add a cardinality constraint $\text{Card}(\ell) \leq 1$, where $\text{Card}(\cdot)$ counts the number of features that are true.

Then, the logical formula defining the component ``get $k$'' is
\begin{align*}
&\mathcal{X} \lor \mathcal{Y},
\end{align*}
where
\begin{align*}
\mathcal{X} &= \ell_k^- \wedge \iota_k^+ \wedge ( \text{Card}(l^+) = 0 ) \wedge ( b^+ = b^- ) \\
\mathcal{Y} &= ( \text{Card}(\iota^-) = 1 ) \wedge \iota_k^+ \wedge \lnot \iota_k^- \wedge (l^+ = l^-) \wedge  \\
&\qquad\left( \forall k_1\;.\; \iota_{k_1}^- \Rightarrow \left( ( b_{k, k_1}^+ = b_{k, k_1}^- - 1 ) \wedge ( b_{\setminus(k, k_1)}^+ = b_{\setminus(k, k_1)}^- ) \right) \right)
\end{align*}
In particular, $\mathcal{X}$ encodes the desired behavior when the agent picks up a loose key $k$, and $\mathcal{Y}$ encodes the desired behavior when the agent unlocks a box to get key $k$.

\section{Experimental Details}

\subsection{Benchmarks}
\label{append:env}

\textbf{2D-craft.}
In this domain, a map is a $10\times10$ grid, where each grid cell is either empty or contains a resource (e.g., wood), obstacle (e.g., water), or workshop. The agent can only observe cells within the distance of 2 units. Since the environment is static, any previously observed cells remain visible. We follow the same approach as in prior work~\cite{psketch} to encode and preprocess the observations: each grid cell is first encoded using a one-hot encoding representing its content (with an entry for unobserved cells); then the preprocessing step extracts the $5\times5$ grid around the current position of the agent as the fine-scale features, and also an aggregated $5\times5$ grid of coarse-scale features which is aggregated over a $25\times25$ region from the original map (after padding) via max pooling.
The flattened version of these features are the inputs to the policy networks in our approach and the baselines. More details can be found in~\cite{psketch} and its code repository.
The test set we use contains tasks with 10 types of goals: get wood, get iron, get grass, get gold, get gem, build plank, build stick, build bridge, build axe, and build ladder. To make the test set more challenging, we include more (15 tasks) from the two hardest goals: get gold and get gem. These goals involve potentially longer horizons to achieve. The rest of the goals are in equal proportion. All our results are averaged over the test set (averaged across different types of goals). This setup follows prior work~\cite{psketch,pagent}. 

For the MLP model architectures, we follow the prior work that originally introduced 2D-craft~\cite{psketch}; in particular, we adopt their model architecture for the actor and critic networks in both our approach and the baselines. We train our hallucinator to operate on state features (e.g. the counts of gems); it takes the state features of the observation as input and predicts the state features of the full map.

\textbf{Box-world.}
In this domain, a map is a $12\times12$ grid with locks and boxes. The agent can only observe cells within the distance of 3 units. As in 2D-craft, since the environment is static, any previously observed cells remain visible. For encoding the observations, each grid cell is encoded using a one-hot encoding representing its content (with an entry for unobserved cells). Following \cite{drrl}, we use a one-layer CNN with 32 kernels of size $3\times3$ to preprocess the map across all approaches before feeding into the policy networks. The test set contains 40 tasks with the number of boxes in the path to the goal varying between 1 to 4; these difficulty levels are in equal proportion.

\textbf{Ant-craft.}
This domain is the same as 2D-craft, except that the agent is replaced with a MuJoCo ant~\cite{ant}, a simulated four-legged robot. We consider a simplified setup where we only model the movements of the ant; the ant directly picks up resources, use tools, and use workshops when it is at the appropriate grid cell (e.g., we do not model the mechanics of grabbing). 

\subsection{Training}
\label{append:details}

We train our models on an NVIDIA GeForce GTX 1080 Ti GPU. The actor-critic training of our approach takes around a day on 2D-craft (400K episodes), 12 hours on box-world (200K episodes), and a day for fine-tuning ant-craft (40K episodes). 
We use the Adam optimizer~\cite{kingma2017adam} with a learning rate of 0.002. We use a batch size of 10 episodes.

\subsection{Ablations}
\label{append:abla}

Here, we provide more detail on the two ablations without a learned hallucinator.

\textbf{Optimistic synthesizer.}
The optimistic synthesizer considers the unobserved parts of the world to be in any possible configuration. If a program can achieve the goal under any one of these configurations, this program is considered to be correct. The optimistic synthesizer chooses the shortest program considered to be correct in this optimistic sense. For example, if the goal of the task is “get gem”, and there is some unobserved grid cells in the current zone, then an optimistic synthesizer will always synthesize the simplest program “get gem”. This baseline also demonstrates the importance of using a hallucinator, instead of a heuristic such as pure optimism.

\textbf{Random hallucinator.}
The random hallucinator randomly predicts the configuration of the unobserved parts of the world. In our experiments, the hallucinator directly predicts the abstract state features, so the random hallucinator simply predicts random values for each entry of the state features (e.g., number of wood in zone 1) under the condition that it does not conflict with existing observations (e.g., predicting number of wood in zone 1 to be 1 when there are already 2 woods observed in zone 1). The purpose of this ablation is to demonstrate the importance of using a learned hallucinator.

\section{Additional Related Work}
\label{append:related}

\textbf{Program synthesis.}
There has been a long line of work on program synthesis, which targets the problem of how to automatically synthesize a program that satisfies a given specification~\cite{deshaw,sketch,flashfill,treeautomata,shivam}. 
More broadly, recent work has explored learning neural network models to predict the program~\cite{robustfill,bunel2018leveraging,chen2018executionguided,chen2018synthesizing,largelm}, as well as using neural models to guide synthesis~\cite{kalyan2018neuralguided,admissible,neuralguided,deepcoder,feng2018program,ellis2019write,chen2020program,nye2020representing,ellis2021dreamcoder}. There has also been work leveraging program synthesis to improve performance in image and natural language domains~\cite{ellis2015unsupervised,ellis2018learning,valkov2018houdini,young2019learning,tian2019learning,huang2020generating,chen2021web}. In contrast, our work uses program synthesis to guide reinforcement learning.

\textbf{Task and motion planning (TAMP).}
TAMP is a hierarchical planning approach that uses high-level task planning and low-level motion planning~\cite{tampnow,inttamp}. TAMP by itself does not handle partial observability; recent work has proposed extensions to address this challenge. For instance, \cite{tamppart} learns a full symbolic program to handle all possible cases---this program tends to be very complex (with many branches) and hence hard to learn. In contrast, our approach learns a simple straight line program that is most likely to solve the task and then replans if needed. Furthermore, \cite{tamppart} only handles discrete partial observations, whereas our approach does not have this restriction. Next, \cite{tampbelief} performs planning in the belief space, which is more similar to our strategy. However, they make the significantly stronger assumption that a structured representation of belief space is available; in particular, they assume a probability distribution over the abstract state space is provided. In general, such a distribution can be difficult to obtain---most deep generative models are unable to explicitly provide the distribution over abstract states; instead, they provide either samples (e.g., GANs and VAEs) or probabilities of given states (e.g., normalizing flows; VAEs can provide a lower bound). As a consequence, it would be difficult to apply this approach to our environments.



\section{Additional Analysis}
\label{append:analysis}

\subsection{Stand-alone evaluations}
\label{append:standalone}

\textbf{Hallucinator.}
We perform additional experiments that measure the prediction accuracy of our trained hallucinator for 2D-craft. We measure accuracy in two ways. The first is the percentage of cases where the predicted state features match the ground truth state features in every entry of the state feature (e.g. the number of zones is an entry, the number of wood in zone 1 is an entry). We call this the “whole” accuracy. The second is the percentage of entries that are correctly predicted, treating each entry of the state feature separately. We call this the “individual” accuracy. We measure accuracy on the test set at different number of steps into the episode. The results are shown in Table \ref{tab:hall}.
As can be seen, the learned hallucinator can correctly predict many entries of the state features, but rarely predicts the whole state features perfectly. This result is due to the intrinsic randomness in the distribution $P(s\mid o)$. Note that accuracy increases with the number of steps into the episodes since the agent has explored more of the map later in the episodes.

\begin{table}[]
    \centering
\caption{Standalone accuracy of the hallucinator}
\label{tab:hall}
    \begin{tabular}{ccc}
\toprule
{\bf Step} & {\bf Whole acc.} & {\bf Individual acc.}\\
\midrule
0 & 0.0\% & 70.9\% \\
20 & 4.5\% & 82.9\% \\
40 & 4.8\% & 85.5\% \\
\bottomrule
\end{tabular}
\end{table}

\textbf{Executor.}
We measure the success rate of the learned executor in our approach at achieving a given component. We evaluate on the test set of 2D-craft environment, focusing on components from the oracle programs. The success rate is 93.8\% (so the failure rate is 6.2\%). The most common failure cases are that the agent gets stuck in some local region of the map. Note that since the program for each task typically includes more than one component, this 6.2\% failure rate will result in $>$6.2\% failure rate in completing the tasks.

\subsection{Baselines trained with fully observed maps}

\begin{figure*}
\centering
\begin{subfigure}[]{0.38\textwidth}
\centering
\includegraphics[width=\textwidth]{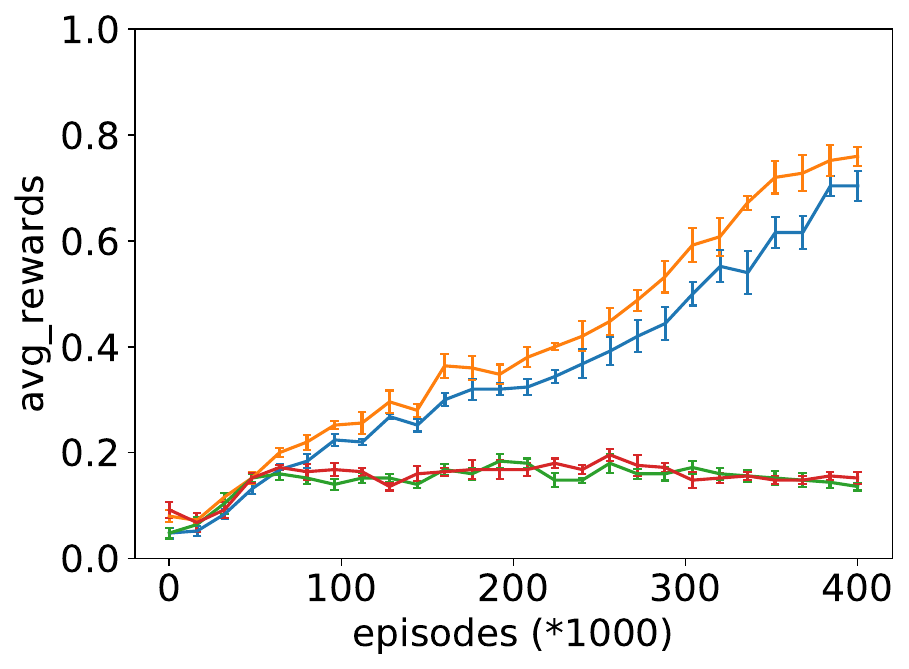}
\caption{}
\label{fig:reward-full}
\end{subfigure}
\begin{subfigure}[]{0.38\textwidth}
\centering
\includegraphics[width=\textwidth]{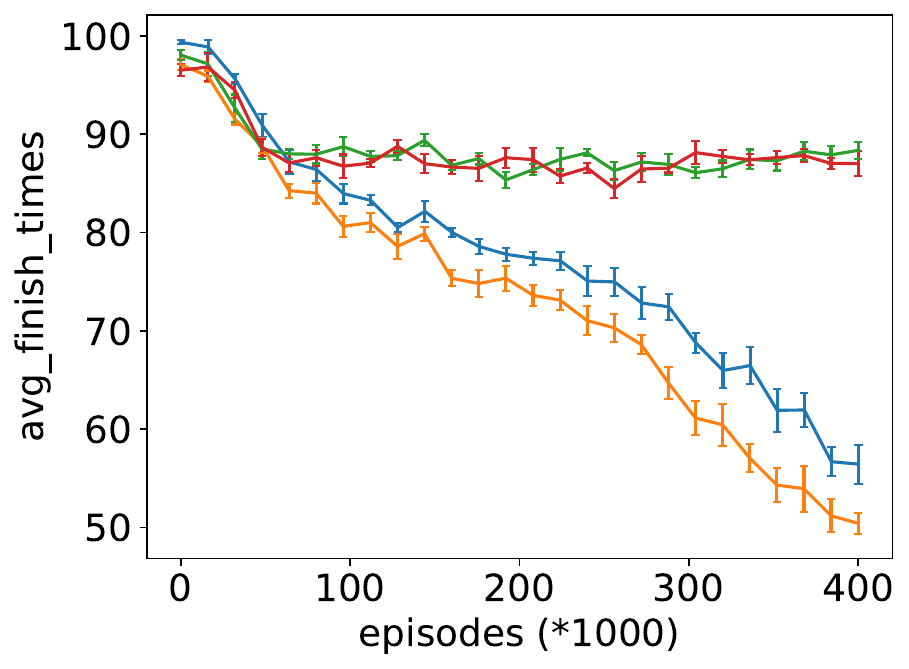}
\caption{}
\label{fig:time-full}
\end{subfigure}
\begin{subfigure}[]{0.15\textwidth}
\centering
\includegraphics[width=\textwidth]{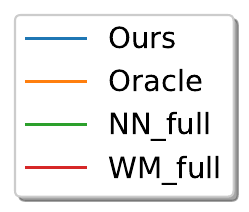}
\vspace{0.5em}
\end{subfigure}
\caption{ Training curves for the 2D-craft environment, comparing our approach with baselines trained on fully observed environments.
(a) The average reward on the test set over the course of training. (b) The average number of steps taken to complete the tasks on the test set. We run all the training with 5 different random seeds, and report the mean and standard error of each metric. We show our approach (``Ours''), program guided agent (``Oracle''), end-to-end neural policy trained on fully observed maps (``NN-full''), and world models trained on fully observed maps (``WM-full'').}
\label{fig:full}
\end{figure*}

In our experiments, we use the programs synthesized from the fully observed maps for training the executor in our approach. This approach avoids repeatedly running the MaxSAT synthesizer during training, which helps speed up training. To ensure this additional information is not responsible for the performance of our approach compared with the non-program-guided baselines, we perform an additional experiment that trains the baselines in fully observed environments. Figure \ref{fig:full} shows results for the 2D-craft environment. As can be seen, our approach continues to significantly outperform the non-program-guided baselines. These results show that providing fully observed map information during training is not the reason our approach outperforms the baselines. 

\subsection{Non deterministic environment}

We perform an additional experiment to study how our approach works when the environment is non-deterministic. We create a non-deterministic version of 2D-craft, where each action has 20\% chance of failing (when a move action fails, the agent move to a random direction; when a use action fails, the action becomes a no-op). Table \ref{tab:nd} shows the results. As can be seen, all the approaches take a longer time to solve tasks in these non-deterministic environments, but our approach continues to significantly outperform the non-program-guided baselines and perform comparably to the oracle.  For the non-program guided baselines, the ratio of test tasks successfully solved does not change significantly, likely because they fail to solve the challenging tasks even when the environment is deterministic. 

\begin{table}[]
\centering
\caption{Performance on the test set for the non-deterministic version of 2D-craft}
\label{tab:nd}
\begin{tabular}{ccc}
\toprule
{} & {\bf Avg. reward} & {\bf Avg. finish step} \\
\midrule
End-to-end & {0.22 (0.02)} & 83.3 (1.8) \\
World models & {0.20 (0.01)} & 83.6 (0.7) \\
\midrule
Ours & {0.47 (0.03)} & 73.1 (1.2) \\
Oracle & {0.50 (0.03)} & {69.9 (1.6)} \\
\bottomrule
\end{tabular}
\end{table}

\section{Additional Examples}
\label{append:additional-examples}

\begin{figure}[h]
\centering
\begin{subfigure}[]{0.24\textwidth}
\centering
\includegraphics[width=\textwidth]{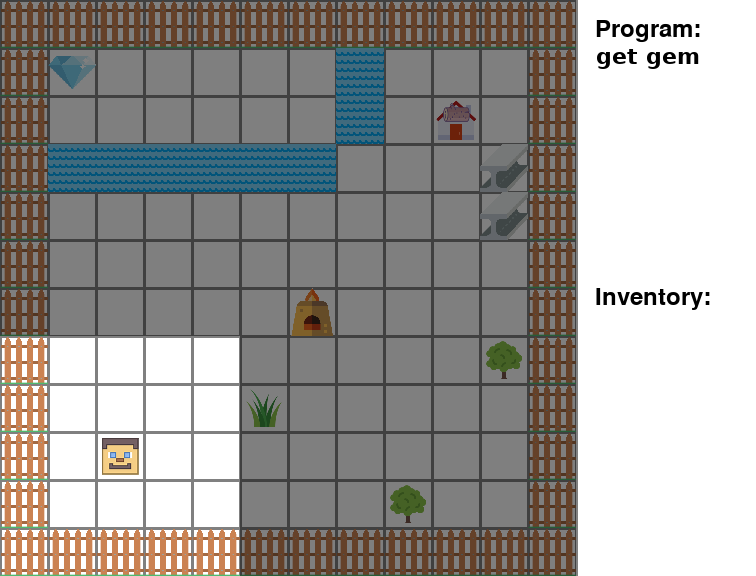}
\caption{}
\end{subfigure}
\begin{subfigure}[]{0.24\textwidth}
\centering
\includegraphics[width=\textwidth]{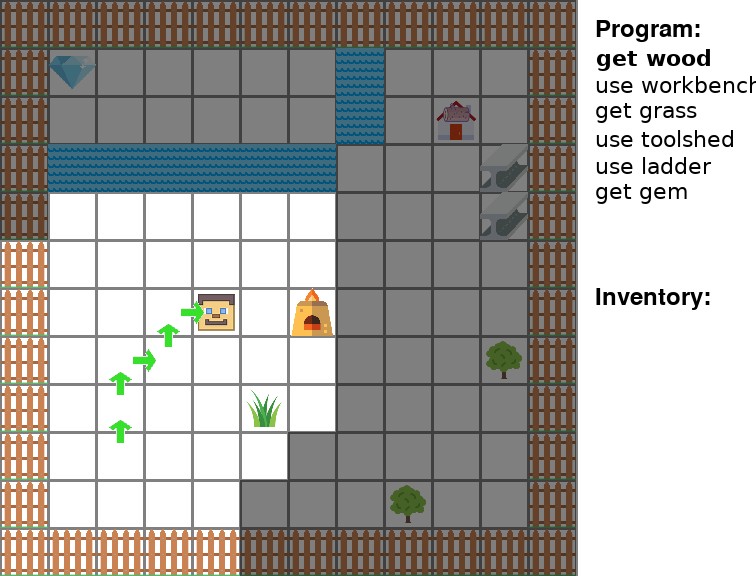}
\caption{}
\end{subfigure}
\begin{subfigure}[]{0.24\textwidth}
\centering
\includegraphics[width=\textwidth]{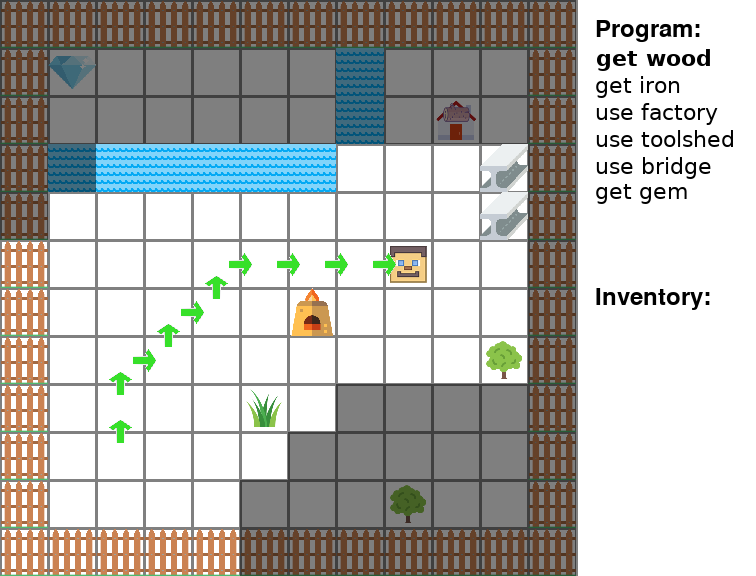}
\caption{}
\end{subfigure}
\begin{subfigure}[]{0.24\textwidth}
\centering
\includegraphics[width=\textwidth]{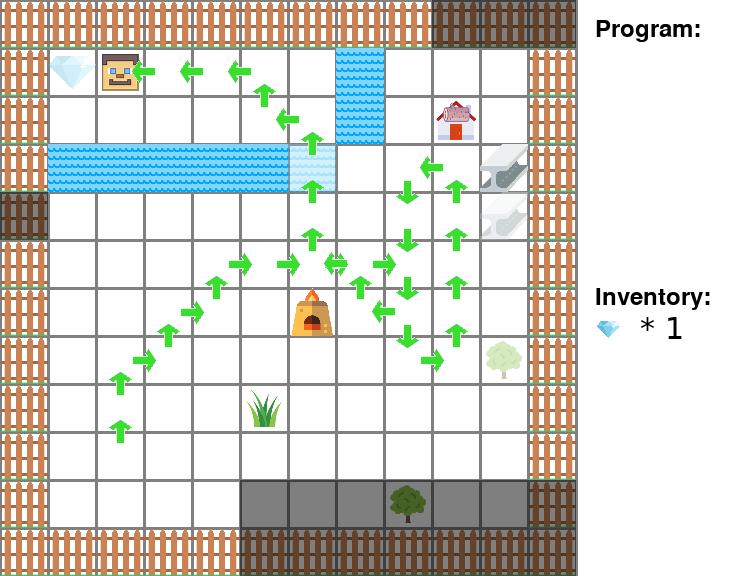}
\caption{}
\end{subfigure}
\caption{Example behavior of our policy in a task with the goal of getting gem. 
(a) The start state. The agent initially hallucinates that there is a gem in the same zone, thus starts with a simple program ``get gem''.
(b) After several steps, the agent observes a grass and a toolshed. Hallucinating based on these new observations, the agent synthesizes a new program that builds a ladder to get gem (which requires grass and toolshed).
(c) After several more steps, the agent observes some water and iron. It re-synthesizes a new program that builds a bridge to cross water. This is a correct program for this task. 
(d) The final state. The agent executes the program and successfully get the gem.}
\end{figure}

\begin{figure}[h]
\centering
\begin{subfigure}[]{0.31\textwidth}
\centering
\includegraphics[width=\textwidth]{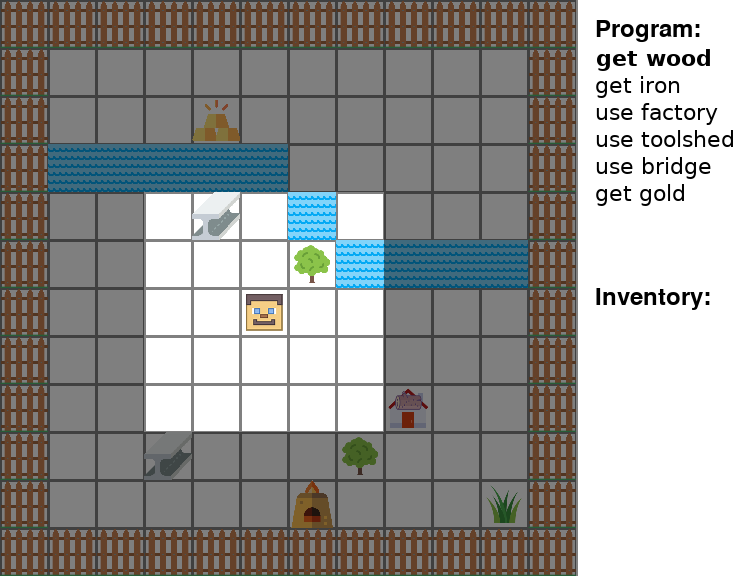}
\caption{}
\end{subfigure}
\begin{subfigure}[]{0.31\textwidth}
\centering
\includegraphics[width=\textwidth]{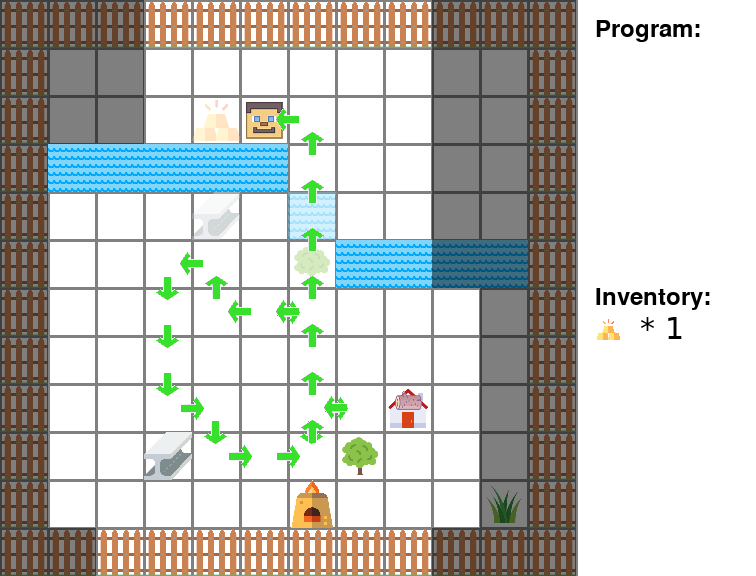}
\caption{}
\end{subfigure}
\caption{Example behavior of our policy in a task with the goal of getting gold. 
(a) The start state. By hallucinating based on the current observations, the agent correctly synthesizes a program that builds and uses a bridge to get to the other zone and get gold. 
(b) The final state.}
\end{figure}

\begin{figure}[h]
\centering
\begin{subfigure}[]{0.31\textwidth}
\centering
\includegraphics[width=\textwidth]{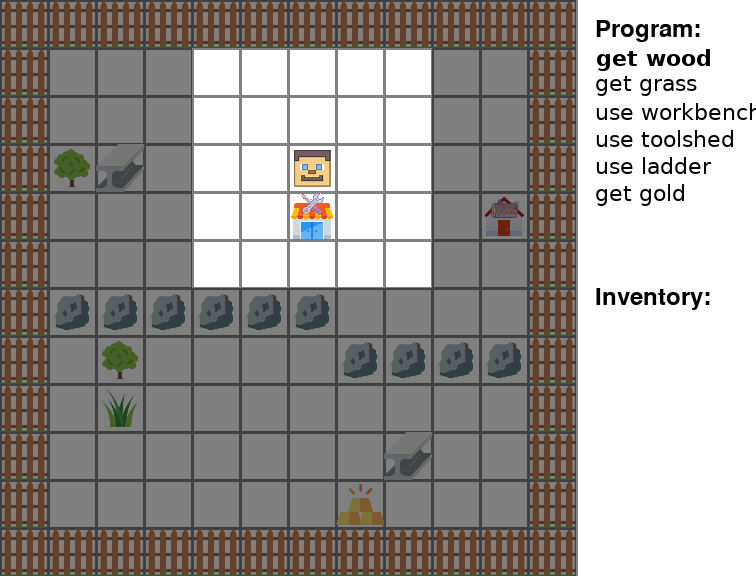}
\caption{}
\end{subfigure}
\begin{subfigure}[]{0.31\textwidth}
\centering
\includegraphics[width=\textwidth]{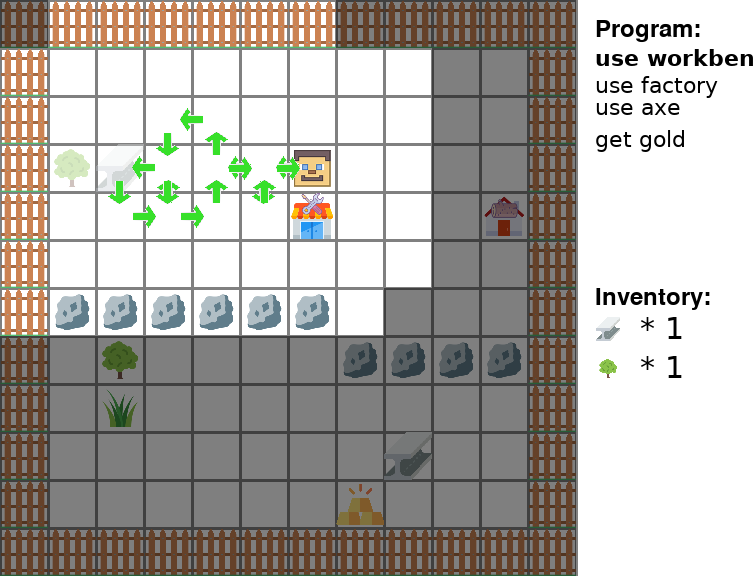}
\caption{}
\end{subfigure}
\begin{subfigure}[]{0.31\textwidth}
\centering
\includegraphics[width=\textwidth]{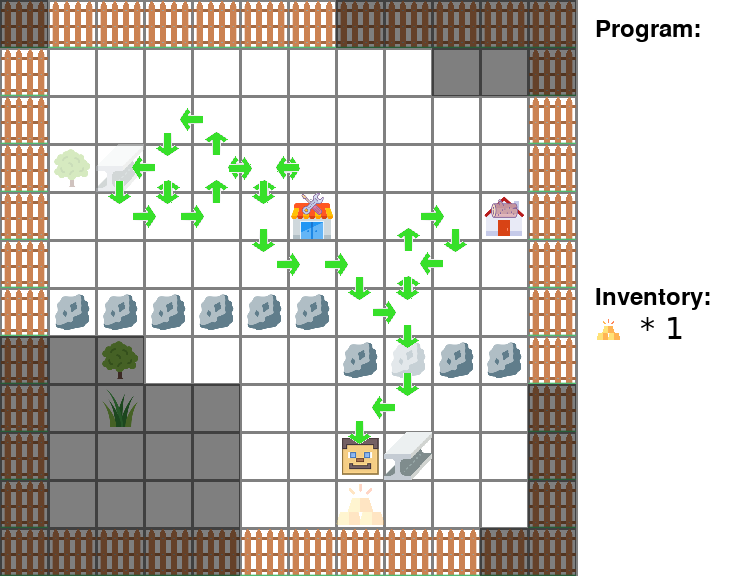}
\caption{}
\end{subfigure}
\caption{Example behavior of our policy in a task with the goal of getting gold. 
(a) The start state. Based on its hallucinations, the agent synthesizes a program that builds and uses a ladder to get a gold in the other zone. However, there is not enough resources and facilities to make a ladder in this map.
(b) The intermediate state when the agent re-synthesizes a new program. With more observations, the agent changes the program to building and using an axe instead, which is a feasible solution in this map. 
(c) The final state.}
\end{figure}

\begin{figure}[h]
\centering
\begin{subfigure}[]{0.24\textwidth}
\centering
\includegraphics[width=\textwidth]{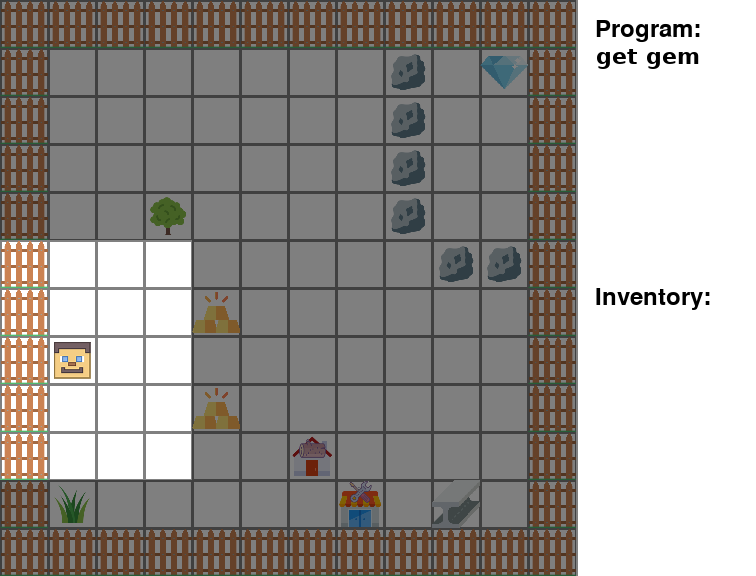}
\caption{}
\end{subfigure}
\begin{subfigure}[]{0.24\textwidth}
\centering
\includegraphics[width=\textwidth]{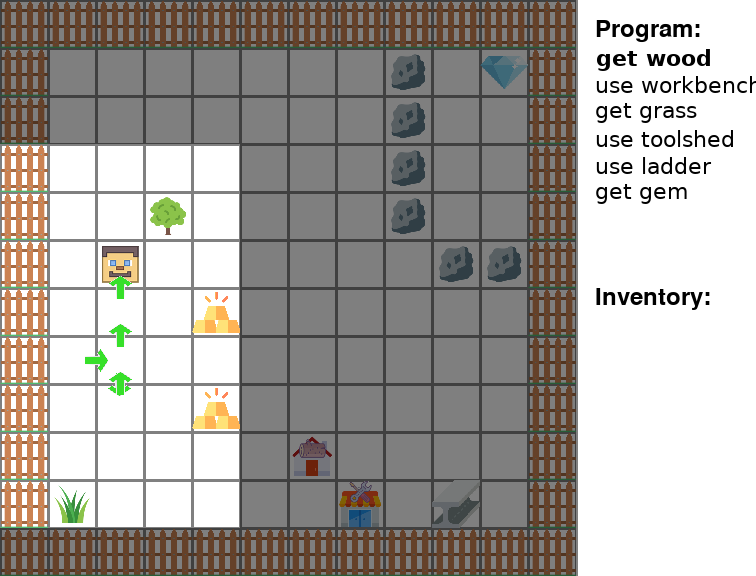}
\caption{}
\end{subfigure}
\begin{subfigure}[]{0.24\textwidth}
\centering
\includegraphics[width=\textwidth]{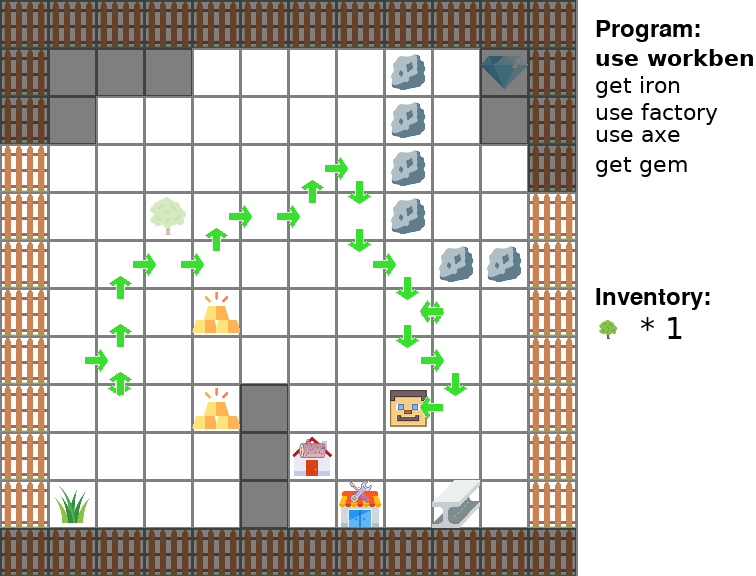}
\caption{}
\end{subfigure}
\begin{subfigure}[]{0.24\textwidth}
\centering
\includegraphics[width=\textwidth]{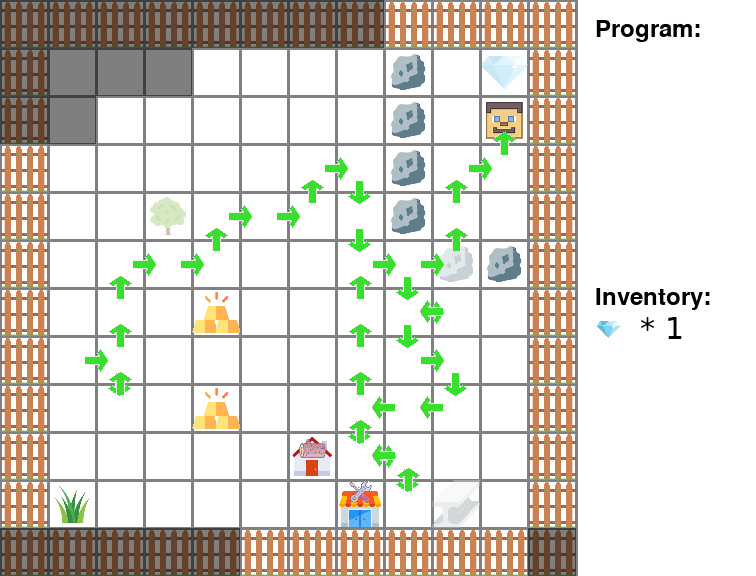}
\caption{}
\end{subfigure}
\caption{Example behavior of our policy in a task with the goal of getting gem. 
(a) The start state. The agent starts with a simple program ``get gem''.
(b) After several steps, the agent observes a grass and a wood. Hallucinating based on these new observations, the agent synthesizes a new program that builds a ladder to get gem (which requires grass and wood).
(c) During its search for workbench, the agent observes all the resources for building an axe. Therefore, it re-synthesizes a new program that builds a axe to cross the stone boundary. This is a correct program for this task. 
(d) The final state. The agent executes the program and successfully get the gem.}
\end{figure}

\end{document}